\documentclass{article}

\usepackage[english]{babel}
\usepackage{PRIMEarxiv}

\usepackage[letterpaper,top=2cm,bottom=2cm,left=3cm,right=3cm,marginparwidth=1.75cm]{geometry}

\usepackage{algorithm}
\usepackage{algpseudocode}

\usepackage{amsmath}
\usepackage{graphicx}
\usepackage[colorlinks=true, allcolors=blue]{hyperref}
\usepackage[numbers]{natbib}
\usepackage{array}
\usepackage{multirow}
\usepackage{booktabs}
\usepackage[table,xcdraw]{xcolor}
\usepackage{arydshln}
\usepackage{booktabs}
\usepackage{pifont}
\usepackage{amssymb}
\usepackage{bbding}
\usepackage{wrapfig}

\definecolor{red}{RGB}{223,74,74}
\definecolor{dark_brown_throughputs}{RGB}{90,63,64}
\definecolor{green}{RGB}{59, 181, 2}
\definecolor{runtime_of_puzzle_chosen_blocks}{RGB}{51,199,145}

\newcommand{\cmark}{\ding{51}}
\newcommand{\xmark}{\ding{55}}

\usepackage{xcolor}
\usepackage{authblk}

\title{Puzzle: Distillation-Based NAS for Inference-Optimized LLMs}

\author{Akhiad Bercovich\textsuperscript{*}}
\author{Tomer Ronen\textsuperscript{*}}
\author{Talor Abramovich}
\author{Nir Ailon}
\author{Nave Assaf}
\author{Mohammad Dabbah}
\author{Ido Galil}
\author{Amnon Geifman}
\author{Yonatan Geifman}
\author{Izhak Golan}
\author{Netanel Haber}
\author{Ehud Karpas}
\author{Roi Koren}
\author{Itay Levy}
\author{Pavlo Molchanov}
\author{Shahar Mor}
\author{Zach Moshe}
\author{Najeeb Nabwani}
\author{Omri Puny}
\author{Ran Rubin}
\author{Itamar Schen}
\author{Ido Shahaf}
\author{Oren Tropp}
\author{Omer Ullman Argov}
\author{Ran Zilberstein}
\author{Ran El-Yaniv}

\affil{\makebox[\linewidth]{\Large NVIDIA}\\
\texttt{\{abercovich, tronen, relyaniv\}@nvidia.com}}

\begin{document}
\maketitle

\begin{abstract} 
Large language models (LLMs) offer remarkable capabilities, yet their high inference costs restrict wider adoption.
While increasing parameter counts improves accuracy, it also broadens the gap between state-of-the-art capabilities and practical deployability. We present \emph{Puzzle}, a hardware-aware framework that accelerates the inference of LLMs while preserving their capabilities.
Using neural architecture search (NAS) at a large-scale, Puzzle optimizes models with tens of billions of parameters.
Our approach utilizes blockwise local knowledge distillation (BLD) for parallel architecture exploration and employs mixed-integer programming for precise constraint optimization.

We showcase our framework’s impact via Llama-3.1-Nemotron-51B-Instruct (Nemotron-51B) and Llama-3.3-Nemotron-49B, two publicly available models derived from Llama-70B-Instruct. Both models achieve a 2.17$\times$ inference throughput speedup, fitting on a single NVIDIA H100 GPU while retaining 98.4\% of the original model's benchmark accuracies. 
These are the most accurate models supporting single H100 GPU inference with large batch sizes, despite training on 45B tokens at most, far fewer than the 15T used to train Llama-70B.
Lastly, we show that lightweight alignment on these derived models allows them to surpass the parent model in specific capabilities.
Our work establishes that powerful LLM models can be optimized for efficient deployment with only negligible loss in quality, underscoring that inference performance, not parameter count alone, should guide model selection.
\end{abstract}

\let\thefootnote\relax\footnotetext{*These authors contributed equally. Other co-authors are listed alphabetically.}

\section{Introduction}
With remarkable advancements in the capability and accuracy of LLMs, these models are increasingly adopted across various domains, ranging from virtual assistants to sophisticated enterprise solutions. 
This adoption trend is accompanied by a growing appetite for larger and more powerful models, driven in part by the aspirational goal of achieving artificial general intelligence (AGI), as evidenced by the industry's push toward increasingly larger-scale LLMs \citep{gpt4, gemini, claude3, llama3} and 
inference-time complexity (GPT-o1, \citep{chain-of-thought, tree-of-thoughts, test-time-training}).
However, the high computational costs associated with these models and their projected future iterations -- particularly during inference -- restrict their accessibility and scalability, thus presenting a significant challenge for widespread commercial applications.

LLMs require a substantial amount of parameters for their training process to converge easily and achieve better generalization \citep{scaling-laws, chinchilla, over-parameterization, double-descent}. This overparameterization not only facilitates optimization, but also provides greater capacity to store knowledge and learn complex patterns across diverse tasks, explaining why larger models consistently demonstrate superior performance \citep{scaling-laws, chinchilla}. However, once trained, many parameters and computations turn out to be redundant for inference, as evidenced by the success of various computational efficiency techniques \citep{no-op_What_Matters_in_Transformers, longformer_window_attention, sink_attention_streaming_llm, sliceGPT, LLM-Pruner, lora, intrinsic-dimensionality}. Yet, LLM architectures remain largely uniform, comprising repeated identical layers, with little consideration given to balancing each block's computational cost against its contribution to overall model predictive performance—a design choice primarily driven by training stability and ease of scaling rather than inference efficiency. This work addresses how to transform a trained LLM from a structure suited for training into one optimized for efficient inference on specific hardware (such as H100), while preserving its accumulated knowledge and predictive performance. Given a ``\emph{parent model}'', our approach explores a large search space of architecture configurations to identify efficient options tailored to meet specific hardware and task-related constraints. This exploration requires a method to reliably estimate the performance of each potential configuration, allowing us to identify models that balance efficiency and accuracy for deployment.

\begin{figure}[h]
    \centering
    \includegraphics[width=\linewidth]{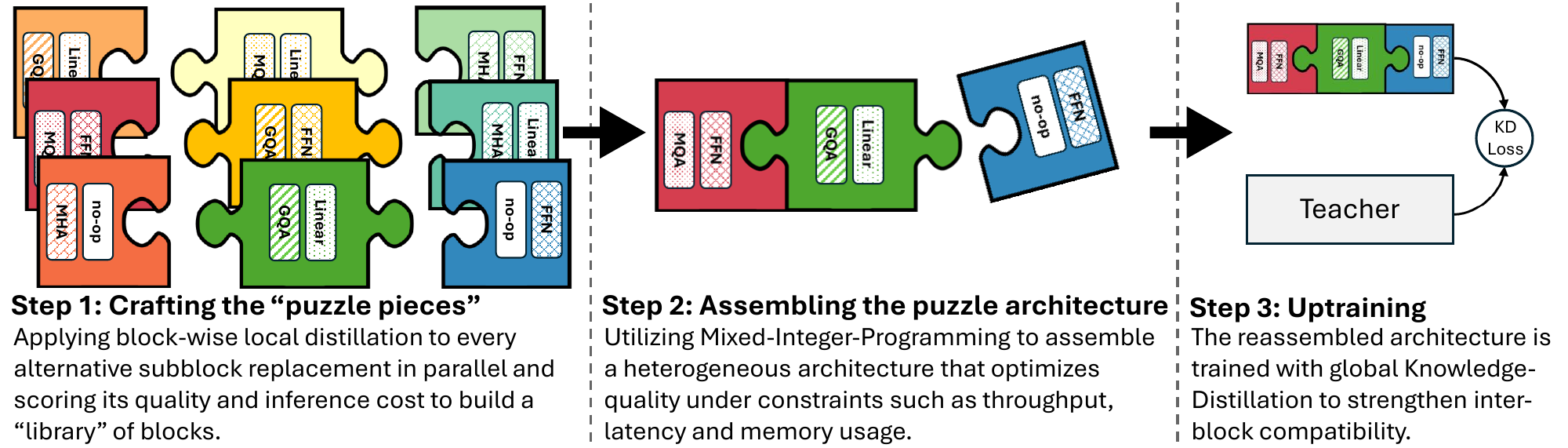}
    \caption{An overview of the three stages of our Puzzle framework.}
    \label{fig:puzzle_overview}
\end{figure}
In this paper we introduce the \emph{Puzzle} framework, summarized in Figure~\ref{fig:puzzle_overview}, which pioneers a decomposed \emph{neural architecture search} (NAS) strategy for LLMs to explore and optimize a vast search space of possible model architectures for target hardware. Inspired by methods in computer vision NAS, and LANA \citep{lana} especially, we define design spaces that include alternative attention and feed-forward network (FFN) layers of varying efficiency degrees, up to a complete layer skip in the extreme case. We then use a \emph{blockwise local distillation} (BLD) (See Section~\ref{sec:bd}) framework to train all these block variants for all layers of the parent LLM in parallel.
Next, we efficiently score each alternative replacement ``puzzle piece'' and search an enormous design space for the most accurate models, while adhering to a set of inference constraints, such as memory size, latency, and throughput. This is done by utilizing a \emph{Mixed-Integer-Programming} (MIP) algorithm.
Lastly, the reassembled model is trained with \emph{Global Knowledge Distillation} (GKD) \citep{hinton-kd}.
Unlike traditional uniform transformer architectures, our NAS framework produces non-uniform models with adapted computation allocation, optimizing each layer's expressivity based on the model’s overall requirements to focus resources where they matter most. This leads to significant gains in efficiency without compromising model expressivity.
By focusing on parent models with SOTA performance, we derive child models pushing the efficient frontier (see Figure~\ref{fig:efficiency_frontier} and Table~\ref{table:ours_8b}), e.g., models which provide the best accuracy per dollar.

Our framework offers several advantages. First,  it enjoys extremely low costs relative to training a model from scratch. For instance, the entire training—BLD before the MIP stage and GKD afterward—required less than 50B tokens to run on Llama-3.1-70B-Instruct, compared to more than 15T tokens used to train the parent model. Even smaller budgets are possible—down to 4B tokens—while still preserving strong performance (see Table~\ref{tab:gkd_efficiency}).
Additionally, our method requires only the parent model’s weights—not its training data—making it ideal for ``open-weights, closed-data'' scenarios where training data of the parent model is not publicly available. This allows practitioners to take any freely-available model and tailor it to their specific hardware or use case.
To demonstrate the effectiveness of our framework, we present Llama-3.1-Nemotron-51B-Instruct (Nemotron-51B), derived from the Llama-3.1-70B-Instruct parent model using Puzzle. Nemotron-51B breaks the efficient frontier of LLMs on a single NVIDIA H100 GPU, establishing a new state-of-the-art for commercial applications, by achieving unmatched throughput and memory efficiency on this hardware. Interestingly, the Nemotron-51B resulting architecture is unique and irregular, with many layers featuring reduced or skipped attention and FFN operations. This design enhances NVIDIA H100 utilization under FP8 quantization while preserving accuracy. We also introduce a derivative of Llama-3.1-8B-Instruct, which breaks the efficient frontier for its throughput slice. While Nemotron-51B also leads within its parameter range, we argue that categorizing models solely by parameter count—such as 50B or 70B—is inadequate for real-world applications. Instead, inference performance under specific hardware, inference engine, quantization levels, budget constraints, and usage profiles—such as varying sequence lengths and batch sizes—should guide model selection.

Conventional approaches to designing LLMs for inference, such as training from scratch or knowledge distillation, present significant challenges. Training a model from scratch is slow and resource-intensive, making it impractical for evaluating multiple configurations.
Knowledge distillation, while generally faster due to guidance from a teacher model, remains prohibitively costly when evaluating multiple candidates from a large search space.
Puzzle circumvents these limitations by conducting architecture search immediately after BLD, during the MIP stage (Stage 2 in Figure~\ref{fig:puzzle_overview}). These stages efficiently identify promising configurations for multiple levels (slices) of inference constraints, without requiring full-model KD for each candidate. The computationally intensive GKD process is reserved for the final stage, after the optimal architectures have been reassembled, enabling Puzzle to focus resources on refining a single optimized model for each slice while keeping the overall cost low.

\textbf{Our contributions}: 

(1) We introduce Puzzle, a framework that applies decomposed neural architecture search to distill an LLM into hardware and inference scenario optimized models. Our framework flexibly supports optimization across ranges of multiple constraints, including throughput, latency and memory usage. Our work pioneers the large-scale use of blockwise distillation and MIP-based architecture search for LLMs, successfully scaling these techniques to tens of billions of parameters while requiring only a fraction of the original training compute. Puzzle is designed to be low-cost, enabling the efficient creation of multiple child models from a single parent LLM—each tailored to different hardware and inference requirements. This scalability makes it feasible to publish optimized variants for diverse use cases and deployment environments.

(2) Using Puzzle, we introduce \textbf{Nemotron-51B} and \textbf{Nemotron-49B}, both optimized for a single H100 GPU, thus setting a new benchmark for commercial applications. We show Nemotron-49B preserves its parent’s 128K context via short uptraining and benefits from RLHF alignment—outperforming Nemotron-51B on some tasks. 
Puzzle's robustness is demonstrated throughout the paper by applying it numerous times with varied constraints, datasets, budgets and parent models.

(3) Our method focuses on optimization for real world scenarios, running on actual inference engines and efficient quantization levels (FP8). We therefore enhance TensorRT-LLM to efficiently support non-uniform blocks and attention mechanisms with varying numbers of key-value heads across layers. This results in models that can directly improve the costs and usability of running LLM inference in practical applications (see Section~\ref{sec:trtllm}).

(4) We provide a comprehensive empirical analysis of the relationship between architectural choices and hardware efficiency, offering insights to guide the design of future hardware-aware LLM architectures.

These contributions advance the field of LLM optimization by transforming powerful language models into efficient, deployment-ready systems while preserving their capabilities. The open-source release of Nemotron-51B demonstrates that state-of-the-art language models can run efficiently on standard hardware, making powerful AI accessible with modest computing resources. This work not only opens new directions for automated architecture optimization but also represents a significant step toward democratizing access to advanced AI technology.

\section{Search Space}
\label{sec:search_space}
The motivation for applying NAS in our work lies in the redundancy found in many trained LLMs. Numerous studies have shown that a significant portion of computations and parameters in LLMs become redundant post-training, during inference \citep{Inheritune, lora, intrinsic-dimensionality}. Prior studies tried to solve these issues with techniques such as pruning \citep{minitron_Compact_Language_Models_via_Pruning_and_Knowledge_Distillation, shortGPT, sliceGPT, LLM-Pruner, sheared-llama, CoFiPruning}, removing entire layers \citep{no-op_What_Matters_in_Transformers, layer_pruning_Ineffectiveness_of_deeper_layers}, local attention methods (e.g., window and sink attention) \citep{longformer_window_attention, sink_attention_streaming_llm}, reducing the number of key-value (KV) heads \citep{mqa, gqa} and many more. Given the high computational and monetary cost associated with running LLMs, optimizing these models to eliminate inefficiencies becomes a critical goal. NAS methods are defined by the search space, the search strategy and evaluation metric of candidate architectures. NAS offers a systematic approach to exploring architectural changes that balances performance and resource constraints, and thus is a prime candidate for reducing inference costs.

In this work, we defined a vast search space, encompassing different operations for each layer of the parent LLM model. 
A \emph{block} is composed of smaller components called \emph{subblocks}. While blocks are user-defined, in LLMs, a block typically refers to a transformer layer, with the subblocks being the attention module and the feed-forward network (FFN).
For each transformer layer $i$, the search space combines options for the attention subblock (denoted $\smash{\mathcal{A}_i = {\{a_j\}}_{j=1}^m}$, where $m$ is the number of possible attention subblocks) and FFN subblock (denoted $\smash{\mathcal{F}_i = {\{f_k\}}_{k=1}^n}$). The attention subblock options could include mechanisms such as standard multi-head attention (MHA), \emph{grouped query attention} (GQA) \citep{gqa} with varying numbers of key-value heads, replacing the attention layer with a single linear layer, or no-op (i.e., entirely skipping the subblock). The FFN options include full or reduced intermediate dimensions (which could be obtained with pruning techniques), linear layers, or no-ops. The combined search space for each transformer layer (or parent block), represented as $\mathcal{A}_i \times \mathcal{F}_i$, captures all possible pairings of attention and FFN configurations.
In this work, each parent transformer layer is replaced by a single corresponding child transformer layer (block), although theoretically, multiple parent layers could be grouped and replaced by a different number of child blocks (i.e., $P$ parent layers to $L$ child blocks). Specifically: 
\begin{itemize}
\item \textbf{Attention subblocks}: For $\mathcal{A}_i$, we consider different variants of GQA with varying numbers of key-value heads (8, 4, 2, and 1). We also include the option to replace the entire attention subblock with a single linear layer or skip it entirely using a no-op operation.
\item \textbf{FFN subblocks}: For $\mathcal{F}_i$, we consider the full intermediate-dimension expansion factor of the parent model, along with reduced dimensions of approximately: 87\%, 75\%, 50\%, 25\%, 20\% and 10\% of the original intermediate size. Furthermore, linear layers and no-op options are also included.
\end{itemize}
These variants offer a tradeoff between memory efficiency, computational cost, and representational power, allowing for flexibility based on specific resource constraints and performance needs. For example, reducing the number of key-value heads (by using GQA with fewer heads) not only speeds up the attention computation, but also helps decrease memory usage by reducing the KV-cache size, which can be crucial for meeting memory constraints or enabling larger batch sizes for better throughput (as GPUs operate more efficiently with larger batches). Using linear layers or reduced FFN dimensions can similarly lower computational requirements, whereas keeping full dimensions maintains higher representational power for better accuracy.
Lastly, in this work, we require that all subblocks within a layer have the same input and output dimensions as their parent. However, a scheme with subblocks of varying embedding dimensions could be developed in future work.

To illustrate the scale of the search space, consider Llama 3.1-70B \citep{llama3}, a model with 80 transformer layers. For the specific instantiation of our framework presented in this work, we defined each transformer layer to have 6 potential alternatives for the attention subblock and 9 alternatives for the FFN subblock, resulting in 54 possible configurations per layer. Consequently, the total number of potential child model architectures is $54^{80}$, which is approximately $10^{138}$ different architectures. This number greatly exceeds the estimated number of atoms in the observable universe ($10^{82}$). Given such an immense number of possibilities, and considering the costs of partially training or even measuring the accuracy of a single LLM architecture, evaluating any representative subset of the search space is computationally infeasible.
Therefore, designing a traditional search strategy and evaluation scheme for NAS in such a vast space is challenging.
To address this, we devised an efficient decomposed local distillation and evaluation framework, and our decomposed search strategy (described in Section~\ref{sec:bd} and Section~\ref{sec:puzzle}). These strategies allow feasible navigation of the search space to find configurations that balance expressivity and efficiency with practical constraints like latency, memory usage, and throughput.

\section{Blockwise Local Distillation}
\label{sec:bd}

\begin{figure}[h]
    \centering
    \includegraphics[width=0.75\linewidth]{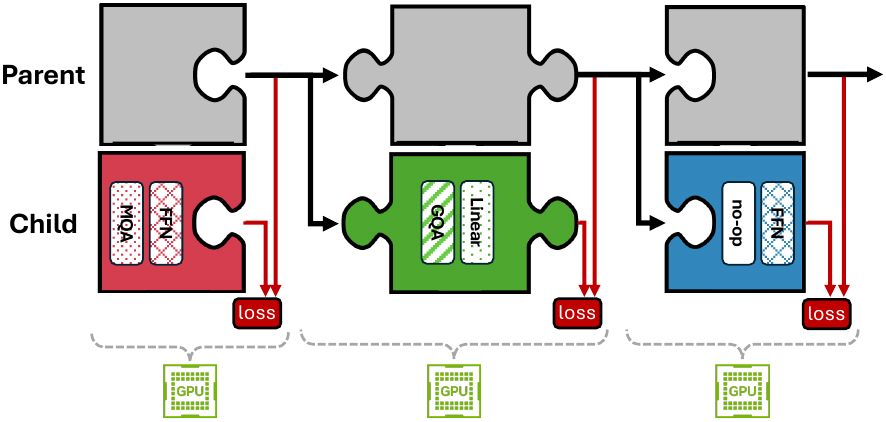}
    \caption{Blockwise local distillation (BLD): each block is trained in parallel and independently.}
    \label{fig:BLD_fig}
\end{figure}

To create capable ``puzzle pieces'' —-a set of trained block variants forming a \emph{block library} for architectural exploration—-we need to set effective weights for each child block. Our method involves efficient, training-free initializations for these blocks (see Section~\ref{subsec:init}). While these initialization techniques are beneficial for low-budget experiments, performance can be significantly improved by distilling knowledge from each parent block to its corresponding child block.
 
Our approach is to decompose the crafting process to operate on individual blocks instead of complete child models, which drastically decreases the computational cost. We train each child block independently and in parallel to locally mimic its corresponding parent block, with only the parent activations being transferred between layers, as illustrated in Figure~\ref{fig:BLD_fig}. This local distillation approach offers several advantages. First, because each child block relies solely on its corresponding parent block, activations and gradients are isolated from other child blocks. This independence enables training blocks separately, leveraging pipeline parallelism across multiple GPUs. Second, each child subblock is trained to mimic a relatively simple function—a single parent subblock—making the process considerably simpler and more stable than training an entire child model. This focused training facilitates faster convergence and allows higher learning rates compared to standard language modeling or GKD methods. Additionally, we find that this approach requires only a small dataset (approximately one billion tokens). Third, each child subblock benefits from high-quality outputs from its preceding parent subblock, rather than the lower-quality outputs typical in global model training, which further enhances convergence speed.

To optimize the performance of each child block, we feed parent activations into the current block and compute a normalized mean squared error (MSE) loss \citep{square_head}. Specifically, we define the loss as \( \mathcal{L} = \frac{\text{MSE}(o_p, o_c)}{\text{MSE}(o_p, 0)} \), where \( o_p \) and \( o_c \) represent the outputs of the original parent block and the modified child block, respectively. 

A primary limitation of BLD is that it does not ensure compatibility between different blocks. This issue arises because each block is trained with inputs from the preceding parent blocks rather than from the outputs of its predecessor blocks within the child model. This prevents later blocks from adapting to the errors of earlier blocks, and may lead to errors propagating and compounding through the child model. To mitigate this, we introduce GKD as a final training phase in our framework (see Section~\ref{sec:global_kd}). Nonetheless, empirical results show that \textbf{the BLD stage alone recovers much of the parent model's performance} (see Table~\ref{table:kd-importance}).

To ensure broad coverage of diverse data domains within limited training schedules, we curated a dataset mixture, termed \emph{Distillation Mix}, for all our distillation training runs. This mixture includes source code repositories, Wikipedia articles, books, news websites, and several other domains. The dataset comprises 224 billion tokens collected from three public datasets: FineWeb \citep{fineweb}, Dolma \citep{dolma}, and Buzz-V1.2 \citep{buzz}. In our BLD experiments, we used 1 billion training tokens. 
We discuss the effect of varying BLD training lengths on downstream tasks in Section~\ref{sec:bld_dataset_size}.

\subsection{Building a Block Library with Decoupled Blockwise Local Distillation}
\label{subsec:build_library}
The first stage of our decomposed NAS framework (further discussed in Section~\ref{sec:puzzle}) is building a ``library'' of trained blocks. In order to cover the entire search space defined in Section~\ref{sec:search_space}, we need to obtain trained weights for each attention variant $a_j$ and each FFN variant $f_k$ in each transformer layer $i$. We consider two methods to train the block library: \emph{coupled BLD} and \emph{decoupled BLD}. For each transformer layer $i$, coupled BLD constructs each possible block variant $[a_j, f_k]_i$ and trains it to emulate the corresponding parent block $[a_\text{parent},f_\text{parent}]_i$. This approach is similar to the aforementioned decomposed NAS methods used in CV. Given the significantly higher computational costs of LLMs compared to CV models, and noting the inherent structure of the transformer layer, we propose decoupled BLD to drastically reduce the cost of building a block library: training $[a_j, f_\text{parent}^{(\text{frozen})}]_i$ and $[a_\text{parent}^{(\text{frozen})},f_k]_i$ separately to emulate $[a_\text{parent},f_\text{parent}]_i$ while freezing the weights of the child subblock that is identical to the parent, and composing the trained subblocks into a full block $[a_j,f_k]_i$ only after training.

\begin{figure}[h]
    \centering
    \includegraphics[width=0.85\linewidth]{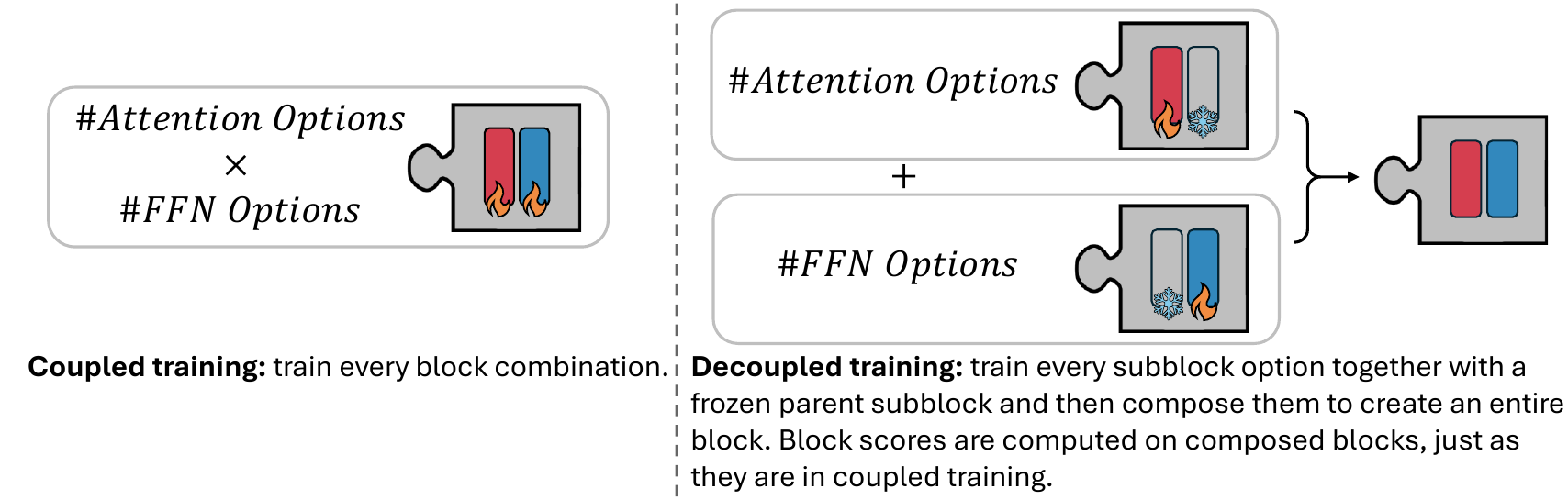}
    \caption{Coupled BLD requires training $|\mathcal{A}_i| \times |\mathcal{F}_i|$ variants per transformer layer, while decoupled BLD requires only $|\mathcal{A}_i| + |\mathcal{F}_i|$ variants per layer, significantly speeding up library construction.}
    \label{fig:coupled_vs_decoupled}
\end{figure}

Given $m$ attention variants, $n$ FFN variants, and $l$ transformer layers, coupled BLD requires training $m \cdot n \cdot l$ variants, while decoupled BLD requires only $(m + n) \cdot l$ variants, significantly speeding up library construction, visualized in Figure~\ref{fig:coupled_vs_decoupled}. This efficiency becomes especially critical with a large toolbox for $\mathcal{A}_i$ and $\mathcal{F}_i$. For instance, with $m = n = 20$ and $l = 80$ layers, decoupled BLD would require training 3,200 blocks compared to 32,000 for coupled BLD. Decoupled BLD enables us to explore a large search space and produce high quality models, without prior knowledge on which combinations of attention and FFN are most preferable. In Section~\ref{subsec:coupling-ablation} we present a technique to combine coupled BLD and decoupled BLD.

\subsection{Initialization of Alternative Subblocks}\label{subsec:init}
To accelerate the distillation process, we introduce training-free initialization techniques for alternative subblocks.
We propose a method to reduce the intermediate dimensions of FFN subblocks by selectively pruning channels based on their contribution. 
Our approach, called \emph{Channel Contribution}, uses an activation-based strategy to estimate channel contribution during forward passes over a calibration dataset, similar to \citep{minitron_Compact_Language_Models_via_Pruning_and_Knowledge_Distillation}.
To guide pruning, we rank intermediate channels based on their impact on the FFN output. This is quantified as the distance between the original FFN output and the output after pruning. The channels with the lowest contributions are prioritized for pruning during subblock initialization.

To formalize our method, let us denote the hidden dimension by~$H$, the FFN intermediate dimension by~$I$, the FFN down projection matrix by~$W^{\text{down}} \in \mathbb{R}^{I \times H}$.
Let $X \in \mathbb{R}^{I}$ represent the FFN's intermediate activations for a single token. The output of the FFN, $Y \in \mathbb{R}^{H}$, is then given by
$$Y = (W^{\text{down}})^\top X = \sum_{k=1}^{I} X_k W^{\text{down}}_{k, :}
$$
We define the per-token contribution of channel~$i$ as:

\begin{align*}
C_i(X) &= \left\| \left( \sum_{k=1}^{I} X_k W^{\text{down}}_{k, :} \right) - \left( \sum_{\substack{k=1 \\ k \ne i}}^{I} X_k W^{\text{down}}_{k, :} \right) \right\|_2 = \lvert X_i \rvert \cdot \left\| W^{\text{down}}_{i, :} \right\|_2  \\
\end{align*}
We then compute the average contribution of each channel across all tokens in the calibration dataset.

When replacing an FFN subblock with a linear layer, we initialize it by computing the product of the up and down projection matrices for FFNs, thereby effectively ignoring the gating mechanism.

In attention subblocks with a reduced number of key-value heads, we initialize the projection matrices for the key and value heads by mean-pooling them into single projection matrices, following the approach used in \cite{gqa}.
When replacing an attention subblock with a linear layer, we use the product of the value and output projection matrices, which simulates the scenario where each token attends only to itself.

\section{Decomposed NAS Search Algorithm for LLMs}
\label{sec:puzzle}

At its core, our decomposed NAS framework is similar to earlier decomposed NAS methods used in computer vision (CV) such as DNA~\citep{dna}, DONNA~\citep{donna}, and especially LANA~\citep{lana}:
\begin{enumerate}
    \item \textbf{Build a block library:} construct a diverse block library using BLD (see Section~\ref{subsec:build_library}).
    \item \textbf{Estimate block resource requirements:} estimate the runtime and memory requirements of each block variant across different scenarios such as inference devices and sequence lengths.
    \item \textbf{Score blocks:} score each block variant in each location inside the network to estimate its quality as an alternative to the parent block in that location.
    \item \textbf{Search architectures:} use a search algorithm to construct ``Puzzle architectures'' that have the best estimated quality under specified runtime and memory constraints.
\end{enumerate}
While these steps mirror established NAS approaches, their application to LLMs presents unique challenges due to the massive scale of the models. Our innovations in scaling these techniques to multi-billion-parameter models while maintaining efficiency are detailed in the following subsections.

\subsection{Estimating Resource Requirements}
\label{subsec:resource_requirements}

Accurate estimation of computational resources is crucial for optimizing LLM architectures for real-world deployment. While theoretical metrics like FLOPs or parameter count are commonly used to approximate computational costs, they often fail to capture the complex realities of hardware acceleration. The actual runtime of neural network operations depends on numerous hardware-specific factors: the number of streaming multiprocessors (SMs), tensor cores, memory bandwidth, and I/O patterns all significantly impact performance. For instance, operations with fewer FLOPs might run slower in practice due to poor hardware utilization, particularly when processing small tensors. This disparity between theoretical and actual performance makes direct measurement on target hardware essential for meaningful optimization.

Memory requirements during inference comprise two distinct components with different scaling behaviors. Parameter memory, while substantial, remains constant regardless of the input size. In contrast, the key-value cache memory scales linearly with both batch size and sequence length, often becoming the dominant factor in long-sequence scenarios. For example, in a model with 32 attention heads and 128-dimensional head size, each token requires 8KB of KV-cache per layer using FP16 precision. For an 8K-token sequence with batch size 64, this amounts to 4GB per layer just for the KV-cache, potentially exceeding the parameter memory.

LLM inference operates in two distinct phases with markedly different performance characteristics. The prefill phase processes the initial input sequence in a single forward pass, enabling high parallelization. While efficient, the computational cost can become significant for long contexts. In contrast, the generation phase processes one query token at a time auto-regressively, requiring repeated forward passes, and often employing paged attention to manage memory efficiently. This difference between phases makes it crucial to actually measure both prefill and generation runtime in the target scenarios (examples measuring Nemotron-51B on several scenarios are shown in Table~\ref{tab:nemotron_throughput}).

Batch size plays a particularly critical role in hardware efficiency, especially during the generation phase. During prefill, even small batches process substantial amounts of data due to sequence length, allowing efficient hardware utilization. However, generation with small batch sizes processes minimal data per layer while still performing all the IO needed to load the layer parameters, leading to severe hardware under-utilization. Increasing batch size creates larger tensors that better utilize GPU resources, often leading to significant improvements in throughput.

Given these complexities, our approach measures resource requirements directly on target hardware across various scenarios. For each block variant, we collect prefill and generation latencies across different  sequence lengths and batch sizes at a chosen quantization level. These measurements directly determine the constraints in our MIP optimization (Section~\ref{subsec:mip}), enabling the search algorithm to find architectures that perform well in actual deployment rather than just in theory.

\subsection{Scoring Architecture Solutions}
\label{subsec:scoring}
A key advantage of decomposed NAS is its ability to estimate the quality of an assembled model based on metrics gathered from its individual blocks. This capability allows search algorithms to explore an enormous search space very efficiently, as the quality of each candidate architecture encountered during the search can be estimated within less than a second, instead of having to actually realize the candidate model and calculate some measure that requires performing forward passes on the entire model such as validation accuracy.
Traditional NAS methods in CV, such as those in \cite{rl_nas} and \cite{evolutionary_nas}, typically rely on full model evaluation, which is computationally prohibitive for LLMs, where both search spaces and inference costs are substantially larger.

We score each block variant at each network location by measuring the impact of replacing only that specific block in the parent model. To do this, we construct a model identical to the parent but with a single block replaced by a trained block from our library, then calculate a performance measure on the entire model. For efficient I/O, when scoring multiple variants, we load onto the GPU only the blocks that differ from the previously evaluated model. We call these scores \emph{replace-1-block scores}. During architecture search, we estimate the quality of a constructed architecture as the sum of its individual replace-1-block scores (see Section~\ref{subsec:mip}). Note that scoring is not performed on candidate models during the search phase -- we estimate the quality of each block only once, then use its replace-1-block score to estimate the quality of any candidate that contains it.

We consider several metrics as potential replace-1-block scores: \emph{(1) downstream accuracy:} accuracy on downstream tasks such as MMLU~\citep{mmlu}. While downstream accuracy is popular in Computer Vision applications, it can be very costly to measure in LLMs, especially given the number of block variants requiring scoring. \emph{(2) LM loss:} causal language modeling loss, defined as the average log-likelihood of a validation corpus under the model's next-token prediction distribution. \emph{(3) KL divergence:} Kullback–Leibler divergence between the next-token prediction distributions of the evaluated model and the parent model, averaged across tokens in a validation corpus. KL divergence is widely used in KD setups as a statistical distance measure but has not been explored for decomposed NAS scoring until now. Our analysis in Section~\ref{subsection:scoring-analysis} highlights the KL divergence effectiveness for block scoring.

\newcommand{\A}{\mathcal{A}}  
\newcommand{\F}{\mathcal{F}}  
\newcommand{\x}{\mathbf{x}}   
\newcommand{\s}{\mathsf{score}} 
\newcommand{\mem}{\mathsf{mem}} 
\newcommand{\rt}{\mathsf{rt}}   

\subsection{Search Algorithm: Mixed-integer Programming}
\label{subsec:mip}

The search space for LLM architecture optimization is enormous (we consider $\sim 10^{138}$ possibilities), as discussed in Section~\ref{sec:search_space}. Efficiently navigating this space requires two key components: a fast method to estimate candidate quality, which we addressed in Section~\ref{subsec:scoring}, and an efficient algorithm to maximize the estimated quality under hardware-specific constraints.

The structure of transformer models naturally frames our optimization problem as a grouped variant of the classical \emph{Knapsack Problem}. Each layer of the model represents a group, containing various block alternatives (attention and FFN variants) as items. Each block alternative has an associated value (its quality score) and multiple costs (parameter memory, KV-cache memory, and runtime characteristics). The objective is to select exactly one block variant from each layer while maximizing the total score and satisfying deployment constraints.

Following \cite{lana}, we formulate this as a Mixed Integer Programming (MIP) problem. Let $x_{i,j}$ be a binary decision variable indicating whether block variant $j$ is selected for layer $i$. For a model with $L$ layers and $K_i$ variants per layer, the optimization problem is:

$$
\begin{aligned}
\text{maximize} \quad & \sum_{i=1}^L \sum_{j=1}^{K_i} \s(i,j)x_{i,j} \\
\text{subject to} \quad & \sum_{i=1}^L \sum_{j=1}^{K_i} [\mem_\mathsf{params}(i,j) + b \cdot \mem_\mathsf{kv}(i,j)]x_{i,j} \leq \mathsf{Memory}_\mathsf{max} \\
& \frac{b \cdot \mathsf{seq\_len}}{\sum_{i=1}^L \sum_{j=1}^{K_i} \mathsf{runtime}(i,j,b)x_{i,j}} \geq \mathsf{Throughput}_\mathsf{min} \\
& \sum_{i=1}^L \sum_{j=1}^{K_i} \mathsf{runtime}(i,j,b)x_{i,j} \leq \mathsf{Latency}_\mathsf{max} \\
& \sum_{j=1}^{K_i} x_{i,j} = 1 \quad \forall i \in \{1,\ldots,L\} \\
& x_{i,j} \in \{0,1\} \quad \forall i,j ~ ,
\end{aligned}
$$

where:
\begin{itemize}
   \item $\s(i,j)$ is the quality score of block variant $j$ in layer $i$, measuring how well it maintains the parent model's performance. If the block scores represent a negative impact (e.g. LM loss or KL divergence) we minimize the sum of scores instead of maximizing it.
   \item $\mem_\mathsf{params}(i,j)$ is the parameter memory required for block variant $j$ in layer $i$, which is shared across all sequences in a batch.
   \item $\mem_\mathsf{kv}(i,j)$ is the key-value cache memory required for a single sequence in layer $i$ with block variant $j$.
   \item $b$ is the batch size - the number of sequences processed in parallel during inference.
   \item $\mathsf{seq\_len}$ is the total sequence length, including both prefill and generation.
   \item $\mathsf{runtime}(i,j,b)$ is the runtime of block variant $j$ in layer $i$ when processing batch size $b$.
   \item $\mathsf{Memory}_\mathsf{max}$ is the maximum allowed total memory, specified to fit the target GPU(s).
   \item $\mathsf{Throughput}_\mathsf{min}$ is the minimum required throughput (tokens per second).
   \item $\mathsf{Latency}_\mathsf{max}$ is the maximum allowed latency per batch.
\end{itemize}

The objective maximizes the sum of quality scores across all selected block variants. The first constraint ensures the total memory usage stays within limits, accounting for both parameter memory (shared across batches) and KV-cache memory (which scales linearly with batch size as each sequence requires its own cache). The second constraint enforces a minimum throughput requirement: for batch size $b$, we process $b \cdot \mathsf{seq\_len}$ tokens within the total runtime, which must meet or exceed $\mathsf{Throughput}_\mathsf{min}$ tokens per second. The third constraint ensures the total processing time for a batch does not exceed the maximum allowed latency. The fourth constraint guarantees exactly one variant is selected for each layer. Note that we do not impose any constraint on the number of model parameters - they are reduced naturally by the search algorithm due to the true scenario constraints.

Since the batch size $b$ is not a variable in this optimization, we solve the MIP problem multiple times with different values of $b$ to explore the runtime-memory trade-off space. Larger batch sizes typically result in higher throughput for all block operations, but also higher latency and more memory for KV-cache storage, which forces a reduction in memory to meet the constraints (such as reducing the number of KV heads). For each set of deployment constraints, we choose the batch size that produced the highest quality architectures. If the target scenario specifies a maximum batch size, such as the typical number of active users for a chat bot, the search can be capped accordingly.

While MIP problems are NP-complete, modern solvers can efficiently handle instances of our size. Using the open-source \texttt{python-mip} package~\citep{python-mip}, we obtain high-quality solutions within seconds. This efficiency enables us to explore multiple architecturally diverse solutions by adding a diversity constraint:

$$
\sum_{i=1}^L \sum_{j=1}^{K_i} x_{i,j}y_{i,j} \leq \alpha \cdot L  \quad  \forall y \in Y ,
$$
where $Y$ is the set of previous solutions and $\alpha \in [0,1]$ controls the maximum allowed similarity. For example, with $\alpha = 0.8$, each new solution must differ from previous solutions in at least 20\% of its layer choices. This constraint helps discover meaningfully different architectures. 

A key feature of our approach is its ability to generate solutions precisely tailored for specific hardware platforms. For example, in platforms with limited memory intended for batch size 1, the algorithm strongly favors memory-saving techniques such as FFN pruning, while de-prioritizing KV-cache optimizations which have minimal impact at batch 1. The hardware specificity extends to architectural features of the inference devices - for example, on H100 GPUs, FP8 quantization offers $\sim$2× acceleration and can be used aggressively, while on A100 GPUs where FP8 is unavailable, different optimization strategies must be employed. Even the difference in inter-GPU bandwidth between H100 PCI-E and NVLink configurations influences the optimal architecture by affecting tensor parallel synchronization costs.

This flexibility enables a powerful ``train once, adapt on demand'' methodology that requires minimal human intervention. After building a block library and computing block scores once, we can efficiently generate different architectures optimized for various deployment scenarios without additional training or manual tuning. The user needs only to specify the available block configurations for each platform - the hardware-specific measurements of block variants naturally guide the optimization toward platform-appropriate solutions. This approach makes Puzzle particularly valuable for real-world deployment, where the same parent model might need to be optimized differently across diverse hardware configurations and deployment constraints.

\section{Post-Puzzle Inter-Block Uptraining}\label{sec:global_kd}
As mentioned in Section~\ref{sec:bd}, the compatibility between adjacent blocks is not accounted for during the BLD step. Individual blocks are not fed with the outputs of their predecessor block in the child model, but rather with the output of the previous corresponding block in the parent model. As a result, after BLD, each block may receive an input distribution different from the one on which it was trained, resulting in sub-optimal performance.
To mitigate this, the last stage in our framework consists of a short end-to-end training of the student using global knowledge distillation (GKD).
GKD, commonly referred to as Knowledge Distillation, is a technique where a ``student'' model is trained to replicate or improve the performance of a ``teacher'' model by learning from its outputs and/or intermediate representations. In the context of this paper, the parent model serves as the teacher and the child model as the student, as the child seeks to inherit the distribution learned by the parent.
Earlier methods \citep{hinton-kd} rely on cross-entropy loss between the teacher and student logits. Later approaches use Kullback-Leibler Divergence (KL Div) loss combined with language modeling loss (i.e. cross-entropy loss on next token prediction), facilitating a faster, more stable, and better performing optimization process \citep{minitron_Compact_Language_Models_via_Pruning_and_Knowledge_Distillation, lu2022knowledge}.

The optimal loss composition depends on the models selected, the training data, and the downstream tasks. We conducted a comprehensive evaluation of various combinations of three types of losses. The \textbf{Language Modeling (LM) Loss} quantifies the divergence between the predicted token probabilities and the target token probabilities in a supervised manner, calculated using the cross-entropy formula:
\begin{equation}
\mathcal{L}_{\text{LM}} = -\sum_{i=1}^{N} y_i \log(\hat{y}_i),
\label{eq:lm_loss}
\end{equation}
where $y_i$ represents the ground truth label for the $i$-th token in the vocabulary, $\hat{y}_i$ is the predicted probability distribution, and $N$ is the size of the vocabulary. 
The \textbf{Cosine Similarity Loss} ensures similarity between the hidden representations of the parent and child models by maximizing the cosine similarity between their respective hidden states. This loss was also used by \cite{minitron_Compact_Language_Models_via_Pruning_and_Knowledge_Distillation}:
\begin{equation}
\mathcal{L}_{\text{cosine}} = \sum_{l=1}^{L} \left( 1 - \frac{\mathbf{h}_c^l \cdot \mathbf{h}_p^l}{\|\mathbf{h}_c^l\| \|\mathbf{h}_p^l\|} \right),
\label{eq:cosine_loss}
\end{equation}
where \(\mathbf{h}_c^l\) and \(\mathbf{h}_p^l\) represent the hidden state vectors from the $l$-th transformer layer of the child and parent models, respectively, with $L$ denoting the total number of transformer layers. While we assume a one-to-one correspondence between teacher and student blocks, this is not a requirement of the framework.
Lastly, the \textbf{Kullback-Leibler Divergence (KLD) Loss} aligns the predicted logits of the child model with those of the parent model to facilitate knowledge transfer. It is computed as:
\begin{equation}
\mathcal{L}_{\text{KLD}} = \sum_{i=1}^{N} p_i \log\left(\frac{p_i}{q_i}\right) ~ ,
\label{eq:KLD}
\end{equation}
where \(p_i\) and \(q_i\) represent the probability distributions derived from the logits of the parent and child models, respectively.

The final loss is the sum of the selected components in each experiment.
All experiments were run on Nemotron-51B, our Llama-3.1-70B-Instruct derivative right after BLD, for approximately 5 billion tokens, which provided sufficient data to observe the relative effects of each loss combination. Our final model was trained on $45B$ tokens using the highest-scoring loss combination: 
\begin{equation}
\mathcal{L}_{\text{GKD}} = \mathcal{L}_{\text{cosine}} + \mathcal{L}_{\text{KLD}} ~ .
\label{eq:GKD}
\end{equation}

As shown in Table~\ref{table:kd-loss-ablation}, LM loss is unnecessary for optimal uptraining and even \textbf{harms} downstream tasks in most cases. This observation aligns with findings from \cite{minitron_Compact_Language_Models_via_Pruning_and_Knowledge_Distillation}, and might result from overfitting to the dataset used during uptraining. Removing the LM loss helps mitigate the inherent data distribution mismatch (open-weights, closed-data) induced by having access only to the parent model, but not to the data on which it was originally trained.
We show the importance of the GKD training stage in Table~\ref{table:kd-importance}.

\begin{table}[!ht]
\centering
\caption{Ablation study for different combinations of LM loss, block (hidden activations) loss, and logits KLD loss. All models (Nemotron-51B, derived from Llama-3.1-70B-Instruct) were  trained for $ \sim5B $ tokens. First row did not undergo uptraining. Adjacent rows with the same color differ only in the $\mathcal{L}_{\text{LM}}$ component. $^*$During the KD process for this combination, the validation $\mathcal{L}_{\text{KLD}}$ consistently increased. $^\dagger$Trained for 45B tokens using $\mathcal{L}_{\text{GKD}}$ defined in Equation~\eqref{eq:GKD}.}
\begin{tabular}{ccccccc}
\toprule
    $\mathcal{L}_{\text{LM}}$~\eqref{eq:lm_loss} & 
    $\mathcal{L}_{\text{cosine}}$~\eqref{eq:cosine_loss} & 
    $\mathcal{L}_{\text{KLD}}$~\eqref{eq:KLD} & 
    \textbf{MMLU} & 
    \textbf{MT-Bench} & 
    \textbf{Average} & 
    \textbf{Validation $\mathcal{L}_{\text{KLD}}$} \\
\midrule
\xmark & \xmark & \xmark & 78.39 & 8.67& 82.55 & 0.19 \\ 
\cmark & \xmark & \xmark & 78.55 & 7.71& 77.83 & 0.31\makebox[0pt][l]{$^*$}\\ 
\rowcolor{gray!10}\cmark & \xmark & \cmark & 79.26 & 8.85& 83.88 & 0.14 \\
\rowcolor{gray!10}\xmark & \xmark & \cmark & 79.33 & 8.68& 83.07 & \textbf{0.10} \\
\cmark & \cmark & \xmark & 79.04 & 7.80& 78.52 & 0.30\makebox[0pt][l]{$^*$}\\
\xmark & \cmark & \xmark & 79.40 & 8.74& 83.40 & 0.16 \\
\rowcolor{gray!10}\cmark & \cmark & \cmark & 79.45 & 8.66& 83.03 & 0.14 \\
\rowcolor{gray!10}\xmark & \cmark & \cmark & \textbf{79.61} & \textbf{8.87} & \textbf{84.16} & 0.11 \\
\midrule
\multicolumn{3}{c}{Llama-3.1-70B-Instruct (parent)} & 81.66 & 8.93 & 85.48 & 0.00 \\
\multicolumn{3}{c}{Nemotron-51B-Instruct (child)$^\dagger$} & 80.20 & 8.99 & 85.10 & 0.08 \\
\bottomrule
\end{tabular}
\label{table:kd-loss-ablation}
\end{table}

\section{Supporting Fast Inference for Variable-Block Architectures in TensorRT-LLM}  
\label{sec:trtllm}
TensorRT-LLM is a highly optimized LLM runtime designed to accelerate inference performance on NVIDIA GPUs. It provides industry-leading performance for both latency and throughput-oriented workloads. The runtime enables LLMs to run on GPUs while utilizing custom paged attention \citep{kwon2023efficientmemorymanagementlarge} kernels to efficiently manage KV caching across sequences in a batch. Furthermore, TensorRT-LLM supports FP8 quantization and various scheduling policies that allow LLM deployments to optimally utilize the underlying hardware. 

To enable the use of Puzzle-generated architectures from the search space (see Section~\ref{sec:search_space}) a major underlying assumption of TensorRT-LLM had to be revised: that all attention layers contain the same number of key and value heads. We devised changes to the paged KV cache strategy that enabled variable GQA ratios within a model. TensorRT-LLM now supports running any architecture from our search space including variable blocks and linear replacements, using FP8 precision for weights, activations and KV cache. Our NAS framework is designed to generate models that run efficiently in real inference scenarios. Therefore full support in inference engines and awareness to runtime considerations when applying on NAS scheme contributes greatly to the usability of resulting models, such as Nemotron-51B.

\section{Main Results}
Using our Puzzle framework, we generated Nemotron-51B as a child derivative of the Llama-3.1-70B-Instruct model. Nemotron-51B achieves a significant improvement in inference efficiency while retaining nearly all the accuracy of its parent model, demonstrating the effectiveness of our approach.

\textbf{Evaluating Model Performance:} To evaluate Puzzle-derived child models like Nemotron-51B, two performance metrics are of primary interest: 
1) \textbf{Accuracy Preservation}: This measures how much of the parent model’s accuracy is retained by the child. A high retention percentage indicates that the Puzzle framework successfully produces high-performing children.
2) \textbf{Computational Efficiency}: This reflects the child model's ability to adhere to the constraints it was optimized for. In our case, the focus is on \emph{throughput}, showing how much we improved the model's suitability for reducing inference cost.

Together, these metrics demonstrate the balance between maintaining model quality and achieving optimization for deployment-specific needs.

\textbf{Accuracy comparison:} Table~\ref{tab:nemotron_accuracy} compares the accuracy of Nemotron-51B with its parent across several benchmarks. On average, Nemotron-51B retains 98.4\% of its parent’s accuracy, even exceeding it on certain benchmarks such as MT-Bench and GSM8K Chat. This underscores the robustness of the Puzzle framework in maintaining model quality despite \textbf{significant} architectural modifications.

\begin{table}[h!]
\centering
\caption{Accuracy comparison of Nemotron-51B with Llama-3.1-70B-Instruct across several benchmarks. Accuracy preserved is the ratio of child to parent accuracy. *Chat prompt as defined in \citet{Nemotron-4-340B}. **version by 
CodeParrot.}
\label{tab:nemotron_accuracy}
\rowcolors{2}{white}{gray!10}
\begin{tabular}{@{}lccc@{}}
\toprule
\textbf{Benchmark}       & \textbf{Llama-3.1-70B-Instruct} & \textbf{Nemotron-51B} & \textbf{Accuracy Preserved (\%)} \\ \midrule
Winogrande \citep{Winogrande}               & 85.08                           & 84.53                 & 99.35                            \\
ARC Challenge \citep{arc}            & 70.39                           & 69.20                 & 98.30                            \\
MMLU \citep{mmlu}                     & 81.66                           & 80.20                 & 98.21                            \\
HellaSwag \citep{hellaswag}                & 86.44                           & 85.58                 & 99.01                            \\
GSM8K \citep{GSM8K}                    & 92.04                           & 91.43                 & 99.34                            \\
TruthfulQA \citep{TruthfulQA}               & 59.86                           & 58.63                 & 97.94                            \\
XLSum English \citep{XL-Sum}            & 33.86                           & 31.61                 & 93.36                            \\
MMLU Chat*                & 81.76                           & 80.58                 & 98.55                            \\
GSM8K Chat*               & 81.58                           & 81.88                 & 100.37                           \\
Instruct HumanEval (n=20) \citep{humaneval}**& 75.85                           & 73.84                 & 97.35                            \\
MT-Bench \citep{MT-bench}                 & 8.93                            & 8.99                  & 100.67                           \\ \bottomrule
\end{tabular}
\end{table}

\begin{figure}[ht!]
    \centering
    \includegraphics[width=0.3\linewidth]{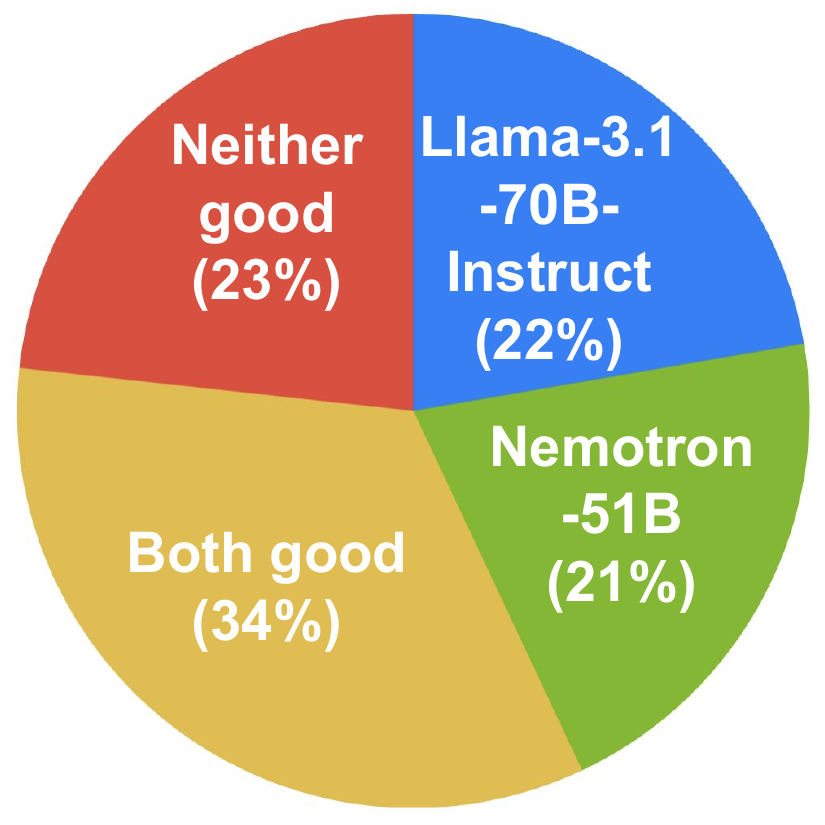}
    \caption{Preference of human annotators in a blind test comparison. Results indicate comparable performance between Llama-3.1-70B-Instruct and Nemotron-51B.}
    \label{fig:human_eval}
\end{figure}

\textbf{Human evaluation:}
We complement the above accuracy performance benchmarks with a human evaluation comparing the parent and child models. A blind test comparison of the models was conducted on a general purpose test set that includes tasks such as reasoning, longform text generation, knowledge and more. Results show comparable performance between the models (Figure~\ref{fig:human_eval}), strengthening the claim that the accuracy degradation is minimal. For more details about the evaluation see Appendix~\ref{sec:human_eval}.

\textbf{Throughput comparison:} Table~\ref{tab:nemotron_throughput} specifies the throughput performance of Nemotron-51B against its parent model across diverse input-output sequence lengths. Nemotron-51B achieves speedups up to 2.17x, enabling larger workloads per GPU and making it highly efficient for deployment.
For each model and hardware configuration, we automatically selected the optimal TP and batch size to maximize throughput per GPU. The inference engine handled this selection dynamically for each run. For example, Nemotron-51B achieved optimal throughput with TP=1 and batch size 256, while Llama-3.1-70B performed best with TP=4 and batch size 384.

\begin{table}[h!]
\centering
\caption{Throughput comparison of Nemotron-51B and Llama-3.1-70B-Instruct across various scenarios. Throughput is measured in tokens per second per GPU (NVIDIA H100). TP\# indicates the number of GPUs used in tensor parallelism. Note: Results were obtained on NVIDIA H100 SXM GPUs with FP8 quantization for weights, activations and KV cache using TensorRT-LLM. Optimal tensor parallelism was used for each model. Input/output sequence lengths indicate the prefill (input) and decode (output) operations performed by the LLM. *TP=1 is not the optimal configuration for Llama-3.1-70B and is included for equal-resource comparison.}
\label{tab:nemotron_throughput}
\rowcolors{2}{white}{gray!10}
\begin{tabular}{@{}lcccc@{}}
\toprule
\textbf{Scenario}        & \textbf{Input/Output} & \textbf{Nemotron-51B (TP\#)} & \textbf{Llama-3.1-70B-Instruct (TP\#)} & \textbf{Speedup} \\ \midrule
Chatbot                  & 128/128              & 5478 (TP1)                  & 2645 (TP1)               & 2.07             \\
Text Generation          & 128/1024            & 6472 (TP1)                  & 2975 (TP4) / 1274 (TP1*)               & \textbf{2.17} / \textbf{5.08}            \\
Long Text Generation     & 128/2048            & 4910 (TP2)                  & 2786 (TP4)               & 1.76             \\
Inference-time compute       & 128/4096            & 3855 (TP2)                  & 1828 (TP4)               & 2.11             \\
Summarization/RAG        & 2048/128            & 653 (TP1)                   & 339 (TP4) / 301 (TP1*)                & 1.92 / 2.17             \\
Stress Test              & 2048/2048           & 2622 (TP2)                  & 1336 (TP4)               & 1.96             \\ \bottomrule
\end{tabular}
\end{table}

\begin{figure}[h!]
\centering
\includegraphics[width=0.6\linewidth]{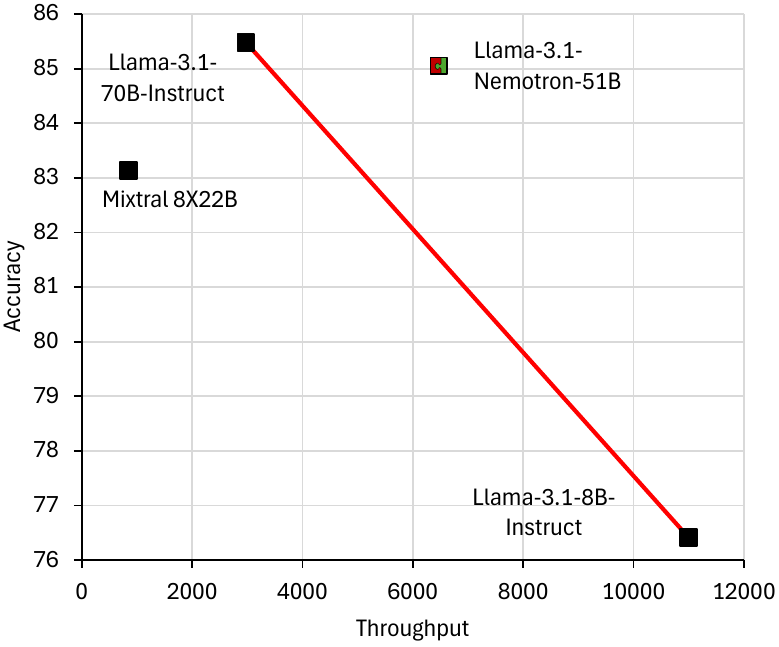}
\caption{Accuracy vs.\@ Throughput performance of Nemotron-51B compared to state-of-the-art models. Throughput is measured on NVIDIA H100 GPUs with optimal TP setting per model, all running in FP8 on a ``text generation'' scenario (see Table~\ref{tab:nemotron_throughput}). The \textcolor{red}{red} line represents the efficient frontier, highlighting models with the best accuracy-to-throughput tradeoff. Accuracy=(MT-Bench $\times$10 + MMLU) / 2}
\label{fig:efficiency_frontier}
\end{figure}

\textbf{Accuracy vs.\@ throughput frontier:} As the field of generative AI evolves, the tradeoff between accuracy and efficiency becomes a key factor in model selection, directly impacting deployment costs. 
Nemotron-51B is designed to achieve this balance, pushing beyond the current efficient frontier. Throughput is inversely proportional to cost, making Nemotron-51B the model with the best accuracy per dollar tradeoff, as shown in Figure~\ref{fig:efficiency_frontier}. To account for both knowledge and conversational capabilities, accuracy is measured as a weighted combination of MMLU and MT-Bench scores: (MT-Bench $\times$ 10 + MMLU) / 2.

\begin{figure}[h]
    \centering
    \includegraphics[width=\linewidth]{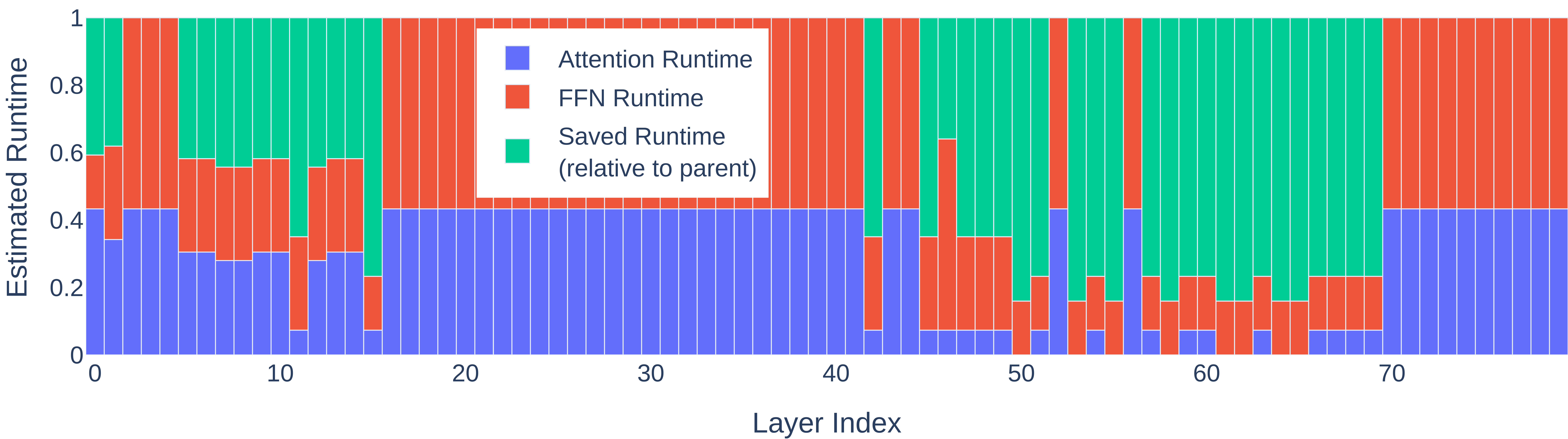}
    \caption{The estimated runtime of the attention and FFN subblocks of Nemotron-51B, relative to the original subblocks of Llama-3.1-70B-Instruct.
    }
    \label{fig:runtime_of_puzzle_chosen_blocks}
\end{figure}

The Nemotron-51B architecture achieves substantial computational savings through strategic reduction of computation across many layers, as shown in Figure~\ref{fig:runtime_of_puzzle_chosen_blocks}. Observing the figure it is evident that our framework discovered significant optimization opportunities in both early layers (0-15) and later layers (45-70), with computed savings shown in \textcolor{runtime_of_puzzle_chosen_blocks}{green}. Different regions exhibit distinct optimization patterns: early layers show balanced reduction in both attention and FFN components, while later layers demonstrate more aggressive attention optimization.
Notably, the framework identifies a central region (layers ~16-42) that maintains full computational capacity, suggesting these middle layers are critical for preserving model performance. This automatically discovered structure demonstrates that large efficiency gains are achievable through careful targeting of computational reduction, while maintaining essential computation where it matters most.

\subsection{Additional Evaluations and Puzzle derivatives}
\textbf{Long-Context Performance:}
We evaluated Nemotron-51B on a subset of the RULER benchmark~\citep{DBLP:journals/corr/abs-2404-06654}. 
Notably, although Nemotron-51B was trained only on sequences up to 8K tokens, it retained over 96\% of its parent’s accuracy 
at 16K tokens, highlighting Puzzle’s ability to preserve performance beyond the direct training length. 
Performance degraded beyond 64K tokens, suggesting that fine tuning on longer contexts 
could extend 
the 
effective context length.

To demonstrate this extension,
we produced \emph{Llama-3.3-Nemotron-49B-Super-Base} (henceforth Nemotron-49B-Base, the base version of Llama-3.3-Nemotron-Super-49B\footnote{\url{https://huggingface.co/nvidia/Llama-3_3-Nemotron-Super-49B-v1}}) from Llama-3.3-70B-Instruct with identical constraints 
to Nemotron-51B and uptrained on an additional 5B tokens at 64K context and 5B at 128K.
Table~\ref{tab:longcontext_49b} shows that Nemotron-49B-Base 
maintains or exceeds its parent’s performance up to 16K tokens, retains over 98\% at 64K, and remains above 94\% at 128K. 
Full performance details are provided in Appendix~\ref{app:full_tables}.

\begin{table}[h!]
\centering
\caption{Comparison of Llama-3.3-70B-Instruct (parent) and Nemotron-49B-Base (child) on RULER for context lengths up to 128K.}
\label{tab:longcontext_49b}
\rowcolors{2}{white}{gray!10}
\resizebox{0.8\linewidth}{!}{%
\begin{tabular}{@{}cccc@{}}
\toprule
\textbf{Context} & \textbf{Parent Average Score} & \textbf{Child Average Score} & \textbf{\textbf{Accuracy Preserved (\%)}} \\
\midrule
4K   & 96.77 & 97.40 & \textbf{100.65}  \\
8K   & 96.46 & 96.59 & \textbf{100.13}  \\
16K  & 95.98 & 96.09 & \textbf{100.11}  \\
32K  & 94.70 & 94.30 & \textbf{99.57}   \\
64K  & 88.91 & 87.39 & \textbf{98.29}   \\
128K & 52.25 & 49.62 & \textbf{94.96}   \\
\bottomrule
\end{tabular}
}
\end{table}

\textbf{Alignment:}
We further aligned \emph{Nemotron-49B-Base} by following the RLHF and instruction-tuning recipe of \citet{HelpSteer2}, 
using 25M tokens. 
Table~\ref{tab:49b_alignment} compares results before and after alignment, showing that Puzzle derivatives can seamlessly undergo RLHF-based alignment 
to boost their accuracy.

\begin{table}[h!]
\centering
\caption{The performance of Nemotron-49B-Base before and after alignment, even surpassing its parent in certain benchmarks.}
\rowcolors{2}{white}{gray!10}
\resizebox{0.7\linewidth}{!}{%
\begin{tabular}{@{}lccc@{}}
\toprule
\textbf{Model}                           & \textbf{MMLU} & \textbf{MT-Bench} & \textbf{Arena Hard \citep{arena_hard}} \\
\midrule
\rowcolor{gray!10}
Nemotron-49B-Base (after lightweight alignment)            & 80.98            & 9.21                 & 82.1                   \\
Nemotron-49B-Base (before alignment)           & 81.03            & 8.77                 & 65.8                   \\
\midrule
\rowcolor{white}
Llama-3.3-70B-Instruct (parent)         & 81.66        & 8.84              & 71.70                   \\
\bottomrule
\end{tabular}
}
\label{tab:49b_alignment}
\end{table}

\textbf{Compact model:} we applied the Puzzle framework to produce a child derivative of Llama-3.1-8B-Instruct, optimized for throughput specifically on an NVIDIA RTX 4090 GPU. This model breaks the efficient frontier for its throughput range, demonstrating Puzzle's ability to deliver highly efficient architectures also on consumer-grade hardware, while preserving the balance between accuracy and performance. Table~\ref{table:ours_8b} highlights this model's superior tradeoff in accuracy and efficiency.

\begin{table}[!h]
\centering
\caption{Accuracy and throughput of our high-throughput child derivative of Llama-3.1-8B-Instruct, which achieves equivalent throughput to Llama-3.2-3B-Instruct and far better accuracy. Throughput is estimated via the sum of measured block runtimes on a single NVIDIA RTX 4090 GPU, measured with an input-output sequence length of 1024 tokens each, the scenario for which this model was optimized. Accuracy = (MT-Bench $\times$10 + MMLU) / 2.}
\rowcolors{2}{white}{gray!10}
\begin{tabular}{@{}lcc@{}}
\toprule
\textbf{Model}                  & \textbf{Throughput*} & \textbf{Accuracy} \\ 
\midrule
Ours (child)                   & 5856                 & 73.98              \\ 
Llama-3.2-3B-Instruct          & 5737                 & 70.36              \\ 
\midrule
\rowcolor{white} Llama-3.1-8B-Instruct (parent)  & 3385                 & 76.40              \\ 
\bottomrule
\end{tabular}
\label{table:ours_8b}
\end{table}

\textbf{Additional Puzzle derivatives:}
Beyond the aforementioned models, Puzzle was also used in the development of the following derivatives:
(1) Puzzle was applied to Llama-3.1-405B-Instruct with constraints requiring a 1.5$\times$ latency speedup and compatibility with a single NVIDIA 8$\times$H100 node (640\,GB) or a single B100 GPU (192\,GB).
The resulting model retained 99.5\% of the parent model's accuracy (averaged across MMLU, MT-Bench, MMLU-Pro, HumanEval, and Arena Hard). 
\emph{FFN Fusion}~\citep{ffn_fusion} was subsequently applied to further improve latency, followed by continued pretraining—resulting in Llama-3.1-Nemotron-Ultra-253B-CPT, which serves as the base for the publicly released Llama-3.1-Nemotron-Ultra-253B \footnote{\url{https://huggingface.co/nvidia/Llama-3_1-Nemotron-Ultra-253B-v1}}.
(2) In \citet{nemotron-h}, a variation of Puzzle—referred to as ``miniPuzzle''—was applied to Nemotron-H-56B-Base under constraints targeting RTX 5090 deployment with a 1M context window. The resulting 47B model retained 99.94\% of the parent model's accuracy (averaged over MMLU, MMLU-Pro, GSM8K, and HellaSwag). The key distinction between miniPuzzle and Puzzle is that miniPuzzle enforces a homogeneous block design across all layers.

\textbf{Training Token Budget:}
The GKD phase described earlier utilized 45B tokens, which might exceed practical budgets for some users. However, in practice, our method can achieve substantial accuracy recovery with significantly fewer tokens. Our experiments, summarized in Table~\ref{tab:gkd_efficiency}, demonstrate that notable accuracy recovery can be realized with reduced token usage. After only 3.7B tokens of GKD, Nemotron-51B recovered 98.8\% of its parent's accuracy on MMLU and MT-Bench benchmarks. Similarly, Nemotron-49B regained 99.63\% of its parent’s accuracy after only 8.68B tokens, and even 98.47\% after just 2.9B tokens. These results indicate that while 45B tokens are modest compared to full model training from scratch, users can flexibly adjust GKD token counts based on available resources without severely compromising accuracy.

\begin{table}[h!]
\centering
\caption{Effectiveness of GKD training with reduced token budgets. The table shows the performance recovery relative to the parent model for Nemotron-51B and Nemotron-49B using various amounts of tokens in GKD.}
\label{tab:gkd_efficiency}
\rowcolors{2}{white}{gray!10}
\resizebox{0.9\linewidth}{!}{%
\begin{tabular}{lccc}
\toprule
\textbf{Model} & \textbf{GKD Tokens (B)} & \textbf{Performance (MMLU / MT-Bench)} & \textbf{Accuracy Preserved (\%)} \\
\midrule
Nemotron-49B-Base & 8.68 & 80.73 / 8.87 & 99.63 \\
Nemotron-49B-Base & 2.9 & 80.72 / 8.675 & 98.47 \\
Nemotron-49B-Base & 0.72 & 80.4 / 8.59 & 97.79 \\
\bottomrule
\end{tabular}%
}
\end{table}

\section{In-Depth Analysis and Ablation Studies}
\label{subsec:ablations}

To better understand the key components and design choices of the Puzzle framework, we conduct a series of detailed analyses and ablation studies. We evaluate the importance of global knowledge distillation, investigate the impact of training dataset size and composition, and analyze how our MIP solver adaptively chooses architectures under varying constraints. These studies not only validate our design decisions but also provide insights into the fundamental trade-offs in LLM architecture optimization and the relative importance of different architectural components.

\subsection{Block Library Construction Ablation Studies}
In Sections \ref{subsec:coupling-ablation} to \ref{subsubsec:limited_search_space}, we analyze the impact of critical decisions in block library construction. These studies examine the fundamental trade-offs between computational cost, dataset characteristics, and model performance. Our key findings are:

\begin{itemize}
    \item \textbf{Coupled vs.\@ Decoupled BLD:} Decoupled BLD reduces training cost by transforming the search space from multiplicative to additive. Combining decoupled BLD for subspace narrowing with coupled BLD for refinement improves accuracy while maintaining computational costs (see Section~\ref{subsec:coupling-ablation}).
    
    \item \textbf{Dataset Composition:} Models trained on the diverse Distillation Mix outperformed those trained on the limited-domain Project Gutenberg, but Gutenberg-trained models still retained ~93\% of performance, showcasing Puzzle's robustness (see Section~\ref{subsubsec:training_dataset_composition}).

    \item \textbf{Training Dataset Size:} BLD achieves strong performance even with smaller token budgets, with diminishing returns beyond 0.5B tokens (see Section~\ref{sec:bld_dataset_size}).

    \item \textbf{Block Scoring Metrics:} KL divergence scoring outperformed LM loss and task-specific downstream scoring, demonstrating better balance between generality and accuracy, although task-specific scoring provides an advantage for customized target tasks (see Section~\ref{subsection:scoring-analysis}).

    \item \textbf{Reduced Search Space:} Constraining the search space to no-op alternatives simplifies optimization and eliminates BLD but results in lower accuracy (75.4 vs.\@ 78.39 MMLU), highlighting the value of diverse block replacements for optimal performance (see Section~\ref{subsubsec:limited_search_space}).
\end{itemize}

\subsubsection{Combining Coupled BLD and Decoupled BLD}
\label{subsec:coupling-ablation}
We investigate the effects of coupling in BLD (see Section~\ref{subsec:build_library}) on Puzzle derivatives of Llama-3.1-8B-Instruct, which has 32 layers. For each layer, the search space we consider contains $|\mathcal{A}_i|=6$ variants of the attention subblock and $|\mathcal{F}_i|=12$ variants of the FFN subblock. With coupled BLD, this would amount to training $6\cdot12\cdot32=2304$ blocks. Decoupled BLD reduces the training requirements to only $(4+10)\cdot32=448$ subblocks, which is considerably more resource-efficient. Note that besides moving from multiplicative composition to additive composition, with decoupled BLD we also do not need to train the no-op and parent variants in $\mathcal{A}_i$ and $\mathcal{F}_i$.

We propose a two-stage technique to reap the benefits of coupled BLD while still keeping the computational cost reasonable. First, we run the full Puzzle framework with decoupled BLD. Then, we analyze the architectures produced by the search algorithm, and identify the most prominent choices of subblock variants. We shrink the search space to include only these choices, then run the full Puzzle framework with coupled BLD on the reduced search space. In our case, we were able to shrink $|\mathcal{A}_i|$ from 6 to 4, and $|\mathcal{F}_i|$ from 12 to 3, resulting in a total number of $4\cdot3\cdot32=384$ blocks to train in coupled BLD, which is similar to the number we trained in decoupled BLD for the larger search space. This approach produced a higher-accuracy architecture. Note, however, that the one created with decoupled BLD was already significantly above the efficient frontier. See results in Table~\ref{table:coupling}. Both models underwent short GKD uptraining.

\begin{table}[!h]
\centering
\caption{The effect of coupled BLD vs decoupled BLD on high-throughput child derivatives of Llama-3.1-8B-Instruct. We found a relevant subspace of the search space using a decoupled BLD Puzzle, then trained coupled BLD on this subspace and ran a separate Puzzle, leading to additional improvement. Throughput is estimated via the sum of measured block runtimes on a single NVIDIA RTX 4090 GPU. Accuracy = (MT-Bench $\times$10 + MMLU) / 2.}
\rowcolors{2}{white}{gray!10}
\begin{tabular}{@{}lcc@{}}
\toprule
\textbf{Model}                  & \multicolumn{1}{l}{\textbf{Throughput*}} & \multicolumn{1}{l}{\textbf{Accuracy}} \\ \midrule
Puzzle with Coupled BLD  & 5856                           & 73.98                        \\
Puzzle with Decouple BLD & 5834                           & 73.10                        \\
Llama-3.2-3B-Instruct  & 5737                           & 70.36                        \\ \bottomrule
\end{tabular}
\label{table:coupling}
\end{table}

\subsubsection{Impact of Dataset Composition on Puzzle-Derived Models}
\label{subsubsec:training_dataset_composition}
We evaluate the robustness of the Puzzle framework to training data composition by comparing performances up through the BLD stage (prior to GKD). For this analysis, we contrast two datasets: our domain-diverse Distillation Mix (described in Section~\ref{sec:bd}) and the English subset of Project Gutenberg \citep{projectgutenberg}. The latter, a dataset predominantly comprising literary works, which lacks diverse coverage of technical, conversational and STEM-specific content, making it an interesting test case for framework robustness.

As shown in Table~\ref{table:data-ablation}, models derived using Project Gutenberg data (for training, pruning and block scoring) demonstrate strong performance despite the dataset's limitations. On general benchmarks like MT-Bench and MMLU, the Gutenberg-trained model achieves 92.7\% and 95.5\% of the performance obtained with Distillation Mix, respectively. Even on STEM categories within MMLU, where the training data's limitations are most relevant, the model maintains 91.7\% of the performance (64.5 vs 70.35).

These results demonstrate that the Puzzle framework can effectively transfer knowledge from the parent model even when the training data provides limited coverage of specific domains. This robustness is particularly noteworthy given that no GKD uptraining was performed, suggesting that our BLD approach effectively preserves model capabilities across domains regardless of the training data composition.

\begin{table}[!h]
\centering
\caption{Benchmark results on Llama-3.1-70B-Instruct derivatives obtained from Puzzle without uptraining applied with different datasets.}
\rowcolors{2}{white}{gray!10}
\begin{tabular}{lcccccccc}
\toprule
  \textbf{Model} & \textbf{MT-Bench}& \textbf{MMLU} & \textbf{MMLU-STEM}\\
\midrule
Gutenberg-Trained & 7.98 & 74.84 & 64.5 \\
DistillationMix-Trained & 8.61 & 78.39 & 70.35 & \\
\bottomrule
\end{tabular}
\vspace{1em}
\label{table:data-ablation}
\end{table}

\subsubsection{Impact of Blockwise Local Distillation Training Dataset Size} \label{sec:bld_dataset_size}

To evaluate the efficiency of BLD under varying training dataset sizes, we conducted experiments using different token budgets. Specifically, we trained the same set of block variants using 0.25B, 0.5B, and 1.0B tokens from the \emph{Distillation Mix} dataset (see Section~\ref{sec:bd}).
After completing the BLD stage for each token budget, we used the MIP optimization stage to generate optimized architectures under similar constraints to those used to produce Nemotron-51B, and evaluated their performance on downstream tasks.

In Table~\ref{table:bld_dataset_size} we present the performance of models trained with different BLD token budgets. Notably, while all models achieved comparable MMLU scores, the MT-Bench results indicate a more pronounced performance improvement as the token budget increases, particularly in multi-turn conversational tasks. However, the improvements diminish as the token budget grows (e.g., the boost from 0.25B to 0.5B tokens is larger than that from 0.5B to 1.0B), suggesting that BLD facilitates rapid recovery of the parent model's performance even with moderate training budgets. These findings imply that longer BLD training can yield better block libraries but with diminishing returns at larger scales.

\begin{table}[h!]
\centering
\caption{Performance comparison of Puzzle-optimized architectures trained with varying BLD token budgets. Metrics include MT-Bench and MMLU scores.}
\label{table:bld_dataset_size}
\rowcolors{2}{white}{gray!10}
\begin{tabular}{lcccc}
\toprule
\textbf{BLD Token Budget} & \textbf{MT-Bench}  & \textbf{MMLU} \\
\midrule
1.0B Tokens               & 8.98                                  & 78.54         \\
0.5B Tokens               & 8.86                                  & 78.44         \\
0.25B Tokens              & 8.51                                   & 78.27         \\
\bottomrule
\end{tabular}
\end{table}

\subsubsection{Impact of Different Block Scoring Metrics}
\label{subsection:scoring-analysis}
\begin{figure}[h!]
    \centering
    \includegraphics[width=0.5\linewidth]{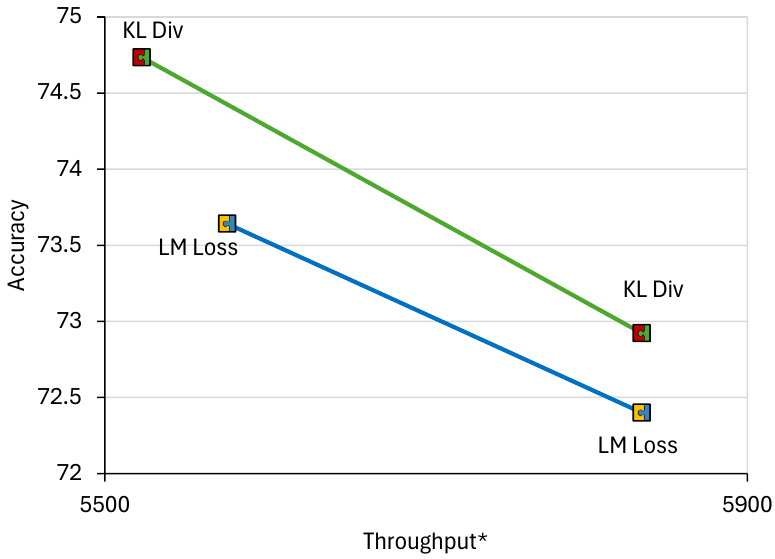}
    \caption{Accuracy vs.\@ Throughput performance of Llama-3.1-8B-Instruct child derivatives, some constructed using LM loss as replace-1-block scores, and some constructed using KL divergence as replace-1-block scores. LM loss aims to capture the general quality degradation induced by changing a specific parent block, while KL divergence aims to capture the distance from the parent model induced by this change. KL divergence results in better architecture choices. Accuracy = (MT-Bench $\times$10 + MMLU) / 2. Throughput is estimated via the sum of measured block runtimes on a single NVIDIA RTX 4090 GPU.}
    \label{fig:lm_loss_vs_kl_div}
\end{figure}

In Section~\ref{subsec:scoring}, we presented three possible metrics for replace-1-block scores: downstream accuracy, LM loss and KL divergence.
We investigate the effects of using different replace-1-block scores when applying the Puzzle framework to Llama-3.1-8B-Instruct with a large search space ($|\mathcal{A}_i|=6, |\mathcal{F}_i|=12$). We compare the use of the model's loss on a validation set (LM loss in our case), a common choice in earlier decomposed NAS methods used in CV, with our proposed method based on KL divergence.
LM loss aims to capture the general quality degradation induced by changing a specific parent block, while KL divergence aims to capture the distance from the parent model induced by this change. As illustrated in Figure~\ref{fig:lm_loss_vs_kl_div}, KL divergence scoring results in better Puzzle architecture choices than scoring with LM loss. All models were constructed using decoupled BLD and underwent short GKD uptraining.

Next, we explore the use of downstream accuracy as replace-1-block scores to customize the algorithm’s block selection for specific target tasks. Our hypothesis is that, within the same budget, different architectures may excel at different tasks because distinct capabilities—such as reasoning, world knowledge, or conversational abilities—are likely concentrated in different parts of the model. Due to the computational expense of calculating downstream accuracy, we use the reduced search space from Section~\ref{subsec:coupling-ablation} to keep the experiment feasible. To evaluate downstream accuracy, we use the MMLU benchmark, splitting its 57 tasks into two nearly equal-sized sets stratified by the MMLU categories \{STEM, Social Sciences, Humanities, Other\}. These subsets are referred to as \emph{Half-MMLU}, with one serving as a ``train'' set for block quality scoring and the other as a ``test'' set for evaluation. Table~\ref{table:half-mmlu} shows that, even with the same library of trained blocks, the block selection process can be customized to construct architectures optimized for specific target tasks. We note that this customization is not without cost: the architecture constructed using Half-MMLU scores achieves the best accuracy on the target task, but its MT-Bench accuracy of 7.57 is lower than the 8.06 MT-Bench accuracy of the architecture constructed according to the more general KL divergence scores. Both models were constructed using decoupled BLD and underwent short GKD uptraining.

\begin{table}[!h]
\centering
\caption{The effect of task-oriented block scoring on high-throughput child derivatives of Llama-3.1-8B-Instruct. We split the tasks in MMLU into two equal-sized sets and use one of them for block quality scoring and the other for evaluation, showing that even with the same library of trained blocks, block selection can be customized to build architectures that fit a desired target task. Throughput is estimated via the sum of measured block runtimes on a single NVIDIA RTX 4090 GPU.}
\rowcolors{2}{white}{gray!10}
\begin{tabular}{@{}lcc@{}}
\toprule
\textbf{Model}                                             & \multicolumn{1}{l}{\textbf{Throughput*}} & \multicolumn{1}{l}{\textbf{Half-MMLU Accuracy (Test Set)}} \\ \midrule
Puzzle: scored with Half-MMLU accuracy (train set)            & 5818                           & 66.24                                       \\
Puzzle: scored with KL divergence & 5834                           & 64.94                                       \\
Llama-3.2-3B-Instruct                             & 5737                           & 60.06                                       \\ \bottomrule
\end{tabular}
\label{table:half-mmlu}
\end{table}

\subsubsection{Effects of Limited Search Space Diversity}
\label{subsubsec:limited_search_space}
To explore the impact of reducing the search space complexity on the Puzzle framework, we constrained alternative child blocks to only allow replacing parent model subblocks with no-op operations. This eliminates the need for BLD as no additional block variants that require training are considered, further reducing the computational costs.

The resulting architecture, optimized using the same MIP approach but limited to no-ops, was evaluated against pre-uptraining Nemotron-51B, which was derived using the same Puzzle pipeline but with a more diverse block variants. As shown in Table~\ref{table:no-op}, the no-op-only model retained high throughput but exhibited a noticeable drop in MMLU accuracy compared to Nemotron-51B (75.4 vs.\@ 78.39).

These results illustrate the flexibility of the Puzzle framework in balancing resource constraints and model performance. Although limiting the search space simplifies the optimization process and reduces training costs even further, it sacrifices the fine-grained architectural customization enabled by a broader range of block alternatives. This demonstrates that a more diverse search space leads to architectures that achieve better accuracy.

\begin{table}[h!] 
\centering 
\caption{Comparison of pre-uptraining Nemotron-51B (derived using the full search space) and a no-op-only variant.} 
\rowcolors{2}{white}{gray!10}
\begin{tabular}{lcc} 
\toprule 
\textbf{Model} & \textbf{MMLU} & \textbf{Throughput (tokens/sec)} \\ \midrule 
Puzzle (No-op only) & 75.40 & 5604.18 \\
Puzzle (Full search space) & 78.39 & 5500.25 \\
\bottomrule 
\end{tabular} 
\label{table:no-op} 
\end{table}

\subsection{Search Algorithm Ablation Studies}
In Sections \ref{subsubsec:MIP_analysis_varying_constraints} to \ref{subsubsec:importance_quality_estimation}, we analyze the MIP optimization process and alternative approaches within the Puzzle framework. These studies focus on how throughput constraints and scoring strategies influence architecture selection and model performance. Our key findings are:

\begin{itemize}
    \item \textbf{MIP Optimization and Throughput Constraints:} The MIP solver adapts architectures to meet various throughput targets, revealing nuanced trade-offs between global and local efficiency. Notably, FFN components are preserved even under strict constraints, highlighting their critical role in maintaining model accuracy (see Section~\ref{subsubsec:MIP_analysis_varying_constraints}).

    \item \textbf{Greedy Algorithm Alternative:} A budget-constrained greedy algorithm, while simpler, resulted in significantly lower accuracy (70.74\% MMLU) compared to MIP-based optimization (78.39\%). This underscores the importance of global optimization for achieving superior accuracy-efficiency trade-offs (see Section~\ref{subsubsec:greedy_search_algorithm}).

    \item \textbf{Data-free Scoring:} Maximizing parameter count as a heuristic produced architectures with sharp performance declines (23.12\% MMLU), emphasizing the necessity of data-driven scoring mechanisms for effective architecture optimization (see Section~\ref{subsubsec:importance_quality_estimation}).

    \item \textbf{Random Architecture Baselines:} Nemotron-51B outperforms both fully random and random-from-library architectures under the same training and speed constraints, underscoring the role of the MIP solver in selecting effective architectural compositions (see Section~\ref{subsubsec:random_architecture_baselines}).
\end{itemize}

\subsubsection{MIP Solution Analysis Under Varying Throughput Constraints}
\label{subsubsec:MIP_analysis_varying_constraints}
\begin{figure}[h]
    \centering
    \includegraphics[width=\linewidth]{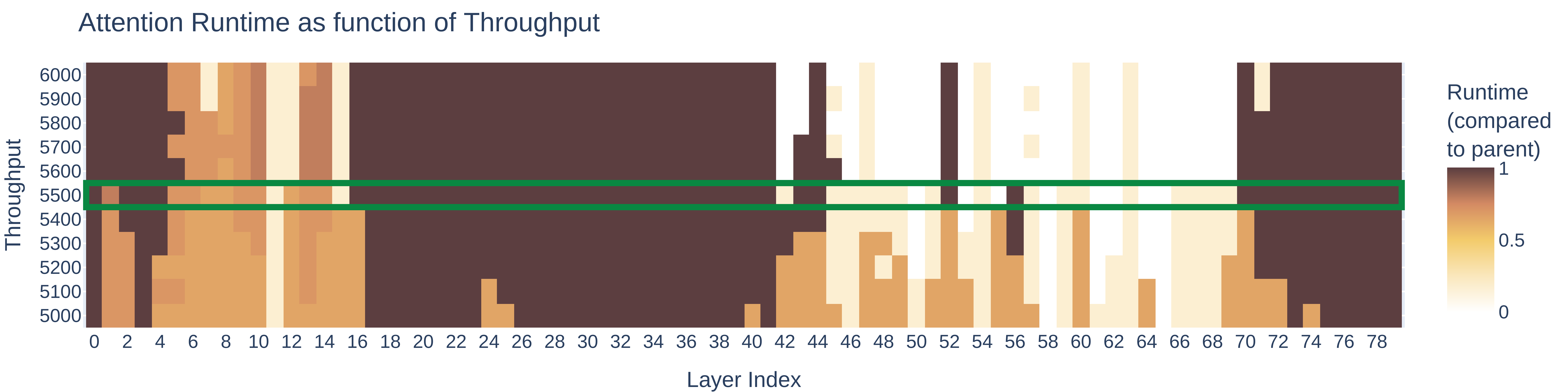}
    \includegraphics[width=\linewidth]{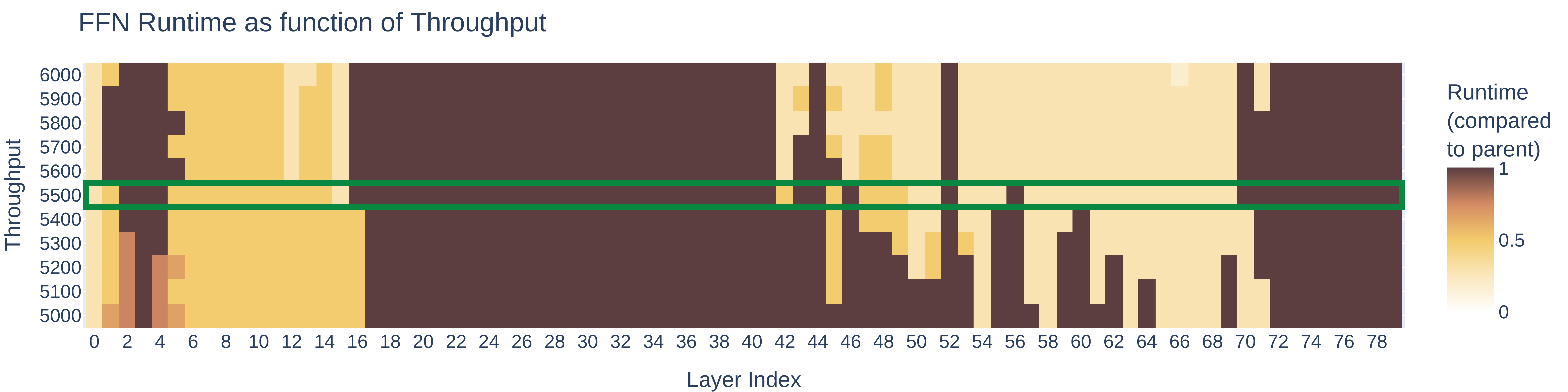}
    \caption{Heatmaps showing how attention and FFN runtime patterns vary with throughput constraints across model layers. \textcolor{dark_brown_throughputs}{Dark colors} indicate higher computational cost relative to the parent model. Each row represents an architecture optimized for a specific throughput target, with Nemotron-51B's configuration (5500 tokens/sec) marked in \textcolor{green}{green}.
    }
    \label{fig:runtime_of_puzzle_chosen_blocks_per_throughput}
\end{figure}

Figure~\ref{fig:runtime_of_puzzle_chosen_blocks_per_throughput} provides an interesting window into how our MIP solver adapts model architecture to different throughput requirements. Each row in the heatmaps represents a distinct architecture optimized for a specific throughput target, with darker colors indicating higher computational cost relative to the parent model. The architecture of Nemotron-51B, corresponding to a throughput target of 5500 tokens per second, is marked in \textcolor{green}{green}. Several intriguing patterns are evident:
\begin{itemize}
\item
\textbf{Counter-intuitive local optimizations:} While stricter throughput constraints generally lead to faster blocks, we observe surprising inversions of this trend. For example, in layers 1-4, the MIP sometimes chooses computationally heavier blocks for higher throughput targets. This counter-intuitive choice suggests that local slowdowns can enable better global optimization, with other layers compensating to meet the overall throughput constraint.
\item
\textbf{Asymmetric treatment of components:} The MIP treats attention and FFN components quite differently. While attention mechanisms are completely eliminated in some layers even under lenient throughput constraints, FFN components are never entirely skipped. This suggests that FFN layers might play a more fundamental role in maintaining model capabilities, while attention mechanisms offer more flexibility for optimization.
\item 
\textbf{Architectural phases:} The heatmap reveals distinct ``phases'' across layer depths. Shallow layers (0-15) show high variability in both attention and FFN, middle layers (16-42) maintain more consistent computation, and deeper layers (43-80) show different patterns for attention versus FFN optimization. This suggests different layers serve distinct roles in the network's information processing.
\item 
\textbf{Throughput-dependent transitions:} The solution patterns show clear transitions as throughput requirements increase, but these transitions are not uniform across layers. Some layers maintain consistent computation across different throughput targets while others show sharp transitions, indicating varying sensitivity to throughput constraints.
\end{itemize}
These patterns demonstrate the sophisticated optimization strategies discovered by the MIP solver, revealing that optimal architectures often require nuanced tradeoffs between local and global efficiency. The preservation of FFN computation even under strict constraints provides empirical evidence for the relative importance of different architectural components in such transformer models.

\subsubsection{Greedy Search Algorithm Alternative}
\label{subsubsec:greedy_search_algorithm}

To further evaluate the impact of the MIP optimization approach, we implemented a budget-constrained greedy search algorithm as an alternative. This algorithm prioritizes simplicity by using a heuristic-based layer-wise selection process, contrasting with the global optimization provided by MIP.

The greedy algorithm operates as follows:

\begin{itemize}
\item At initialization, the runtime and memory budgets which are derived from the required constraints, are split equally across layers. 
\item  Layer scoring: Each layer is assigned a score based on a heuristic metric designed to estimate how easily it can be replaced with minimal performance degradation. In our implementation, this metric was the mean replace-1-block KL divergence score across all block variants for the layer. Layers with lower average scores were deemed easier to optimize.
\item Sequential replacement: Layers are processed in ascending order of their scores. For each layer, the algorithm selects the block variant with the lowest replace-1-block KL divergence score that satisfies the layer's runtime and memory budget.
\item Constraint adjustment: After selecting a block for the current layer, the remaining runtime and memory savings are added to the next layer's budget. This allows for a more dynamic algorithm that allocates more resources for layers that are harder to replace.
\end{itemize}

Table~\ref{table:greedy-alg} compares the performance of the greedy algorithm with the MIP-derived pre-uptraining Nemotron-51B. The results show that the greedy algorithm leads to a significant drop in accuracy, highlighting the importance of global optimization. Specifically, the model derived using the greedy algorithm achieves a substantially lower MMLU accuracy of 70.74\%, compared to 78.39\% for the MIP-derived model, despite both architectures meeting the same throughput constraint.

\begin{table}[h!] 
\centering 
\caption{Comparison of the budget-constrained greedy algorithm and MIP as search algorithms for Puzzle. Results are shown for pre-uptraining Nemotron-51B under identical throughput constraints.} 
\rowcolors{2}{white}{gray!10}
\begin{tabular}{lcc} 
\toprule 
\textbf{Optimization Method} & \textbf{MMLU} & \textbf{Throughput (tokens/sec)} \\
\midrule Greedy Algorithm & 70.74 & 5500.30 \\
MIP & 78.39 & 5500.25 \\
\bottomrule 
\end{tabular} 
\label{table:greedy-alg} 
\end{table}

These findings underscore the critical role of global optimization in the Puzzle framework. By considering jointly optimizing block selection, MIP achieves a significantly better balance between efficiency and accuracy, making it indispensable for extracting the full potential of the Puzzle framework.

\subsubsection{Importance of Data-Driven Quality Estimation in Architecture Scoring}
\label{subsubsec:importance_quality_estimation}
To further understand the importance of quality estimation in search space optimization, we conducted an experiment using a simple heuristic: scoring blocks based on their parameter count. While not a data-driven approach, parameter count serves as a straightforward metric often associated with improved performance in LLMs. This experiment aimed at assessing whether such a basic metric could provide sufficient guidance in the search space, shedding light on the role of more nuanced scoring mechanisms.

Under this method, the search algorithm is simplified to selecting the block variant with the largest number of parameters that satisfied throughput and memory constraints. This resulted in an architecture composed uniformly of high-parameter blocks across all layers, without considering layer-specific block quality and representational needs.

As shown in Table~\ref{table:max-params}, maximizing parameter count leads to a sharp drop in MMLU accuracy compared to the architecture derived using the full Puzzle framework with MIP optimization. Despite having similar throughput, the simplistic parameter-maximization approach fails to achieve competitive results, underscoring the necessity of quality-aware block scoring, and different layers require different computational budgets for optimal performance.

\begin{table}[h!] 
\centering 
\caption{Comparison of maximizing parameter count with Puzzle's MIP-based optimization as search algorithms. Results are shown for pre-uptraining Nemotron-51B under identical throughput constraints.} 
\rowcolors{2}{white}{gray!10}
\begin{tabular}{lcc} 
\toprule 
\textbf{Optimization Method} & \textbf{MMLU} & \textbf{Throughput (tokens/sec)} \\
\midrule Maximizing Parameters & 23.12 & 5727.08 \\
pre-uptraining Nemotron-51B & 78.39 & 5500.25 \\
\bottomrule 
\end{tabular} 
\label{table:max-params} 
\end{table}

\subsubsection{Evaluating Random Architectures as Search Baselines}
\label{subsubsec:random_architecture_baselines}

To further validate the effectiveness of our search algorithm, we evaluated several baselines where block selection was performed randomly. All models were trained with 10B tokens.

The first baseline (``Random-from-block-library'') randomly samples block variants from the full block library while satisfying the same throughput constraint as Nemotron-51B, ignoring their block scores. The second (``Fully Random'') uses a completely random architecture that is not constrained to trained block variants, but adheres to the same speed constraints. Finally, we added a third baseline—\textit{Parent-Randomized}—which evaluates Llama-3.1-70B with randomized weights but no architectural modifications.

As shown in Table~\ref{table:random_baselines}, Puzzle significantly outperforms all baselines. Notably, the random-from-library baseline achieves only 86.6\% of Nemotron-51B’s performance, despite being constructed from the same trained blocks. This experiment further emphasizes the role of MIP in selecting high-quality architectural compositions from the library.

\begin{table}[h!]
\centering
\caption{Comparison of Nemotron-51B with randomly constructed architectures, all trained for 10B tokens. The random-from-library variant uses trained blocks, while the fully random variant ignores the block library. The Parent-Randomized model uses Llama-70B’s architecture with random weights.}
\label{table:random_baselines}
\rowcolors{2}{white}{gray!10}
\resizebox{\linewidth}{!}{%
\begin{tabular}{lcccc}
\toprule
\textbf{Model} & \textbf{MMLU} & \textbf{MT-Bench} & \textbf{Avg. Accuracy} & \textbf{Relative to Llama-70B} \\
\midrule
Nemotron-51B (10B tokens) & 79.7 & 8.89 & 84.30 & 98.61\% \\
Random-from-block-library & 66.02 & 8.20 & 74.01 & 86.58\% \\
Fully Random & 23.13 & 0.89 & 16.02 & 18.73\% \\
Parent-Randomized & 23.42 & 0.95 & 16.46 & 19.25\% \\
Llama-3.1-70B & 81.66 & 8.93 & 85.48 & 100\% \\
\bottomrule
\end{tabular}
}
\end{table}

\subsection{Global Knowledge Distillation Uptraining Importance}
We assess the significance of the final GKD uptraining phase, in enhancing the accuracy of child models derived from Llama-3.1-70B-Instruct and Llama-3.1-8B-Instruct. The results presented in Table~\ref{table:kd-importance} demonstrate that this stage contributes to improvements in both the MMLU and MT-Bench benchmark scores.

\begin{table}[!ht]
\centering
\caption{Impact of global knowledge distillation uptraining on MMLU and MT-Bench benchmark scores for child models derived from Llama-3.1-70B-Instruct and Llama-3.1-8B-Instruct.}
\resizebox{0.85\textwidth}{!}{
\begin{tabular}{@{}lcccc@{}}
\toprule
    \textbf{Model Name} &
    \textbf{GKD Uptraining} &
    \textbf{MMLU} &
    \textbf{MT-Bench} &
    \textbf{Average} \\ 
\midrule
Llama-3.1-70B-Instruct (parent)& - & 81.66 & 8.93 & 85.48 \\
\rowcolor{gray!10}& \xmark & 78.39 & 8.67  & 82.55  \\
\rowcolor{gray!10}\multirow{-2}{*}{Nemotron-51B-Instruct (child)}   & \cmark & 80.20 & 8.99  & 85.10  \\ \midrule
Llama-3.1-8B-Instruct (parent)& - & 69.40 & 8.34 & 76.40 \\
\rowcolor{gray!10}   & \xmark & 65.25 & 7.29 & 69.06  \\
\rowcolor{gray!10}\multirow{-2}{*}{Child derivative of Llama-3.1-8B-Instruct (child)} & \cmark & 65.46 & 8.25 & 73.98  \\
\bottomrule
\end{tabular}
}
\label{table:kd-importance}
\end{table}

\subsection{Comparison with Related Work} \label{subsubsec:related_techniques}

There are many different techniques to prune or compress models, most of which could be complementary to Puzzle. Some pruning methods, for example, could be used to construct alternative blocks for Puzzle's block library. Nevertheless, in this section we compare several established methods directly against the basic Puzzle framework, without integrating them.

We compare Puzzle against structured sparsity (Wanda \citep{sun2023wanda}) and low-rank approximation (similar to \citep{khodak2024initialization} with subsequent distillation) applied to Llama-3.1-70B under similar throughput constraints. Wanda applied 2:4 structured sparsity without additional training, while the low-rank method used factorized layers followed by distillation. Nemotron-51B significantly outperformed both methods, achieving 99.49\% of the parent's average accuracy (MMLU and MT-Bench), compared to 92.23\% for Wanda and 88.96\% for low-rank approximation (see Table~\ref{table:wanda_low_rank}). Subsequent distillation post-pruning with Wanda slightly improved MMLU (73.69) without impacting MT-Bench. Moreover, since both structured sparsity and low-rank approximations represent subsets of Puzzle's broader search space, integrating these approaches into Puzzle could further enhance performance.

\begin{table}[h!]
\centering
\caption{Comparison of Puzzle, Wanda (structured sparsity), and low-rank approximation methods on Llama-3.1-70B derivatives under similar throughput constraints.}
\label{table:wanda_low_rank}
\rowcolors{2}{white}{gray!10}
\resizebox{\linewidth}{!}{%
\begin{tabular}{lcccc}
\toprule
\textbf{Model} & \textbf{MMLU} & \textbf{MT-Bench} & \textbf{Average Accuracy} & \textbf{Accuracy Preserved (\%)} \\
\midrule
Nemotron-51B & 80.20 & 8.99 & 85.05 & 99.49\% \\
Wanda \citep{sun2023wanda} & 72.99 & 8.39 & 78.44 & 92.23\% \\
Low-rank & 72.87 & 8.01 & 76.05 & 88.96\% \\
Llama-3.1-70B (Parent) & 81.66 & 8.93 & 85.48 & 100\% \\
\bottomrule
\end{tabular}
}
\end{table}

Other methods such as Minitron \citep{minitron_Compact_Language_Models_via_Pruning_and_Knowledge_Distillation}, ShortGPT \citep{shortGPT}, and SlimGPT \citep{slimgpt} share conceptual similarities with Puzzle, but each represents a constrained subset of Puzzle's broader optimization space. For the most part, Minitron restricts modifications to homogeneous block replacements across all layers, ShortGPT focuses exclusively on redundant layer removal, and SlimGPT employs incremental pruning ratios via a fixed heuristic. Puzzle generalizes and extends these approaches, allowing for heterogeneous, layer-specific modifications, diverse block alternatives including no-op layers, and customizable pruning ratios optimized globally through MIP-based optimization.

\section{Conclusions and Future Directions}

In this work we present Puzzle, a framework that modifies LLMs from their over-parameterized training configurations to optimized, inference-efficient architectures tailored for specific hardware. 
Puzzle achieves these improvements with remarkable efficiency in training resources. Requiring fewer than 50B tokens—compared to the trillions needed to train models from scratch—Puzzle produces high-performing models at a fraction of the usual cost. This extreme search and training efficiency still results in drastic inference performance improvements of Puzzle optimized 
models, thus enabling widespread and efficient deployment of LLMs across data centers and edge devices.

The success of Puzzle opens several promising directions for future research. Our introduction of decoupled BLD, which is significantly more efficient than coupled BLD, makes it feasible to evaluate a much larger set of potential blocks within a single optimization run. This efficiency enables exploration of novel operations as alternative Puzzle blocks, such as variable window attention mechanisms \citep{longformer_window_attention}, \emph{state-space models} \citep{ssms, mamba}, or other architectural innovations. The framework could also be extended to optimize models for specific capabilities, such as Chain-of-Thought reasoning or multimodal tasks, including vision-language models 
\citep{llava, transfusion} and retrieval-augmented generation \citep{rag}.
\citep{llava} and retrieval-augmented generation \citep{rag}.

\section*{Acknowledgments}
We thank Saurav Muralidharan and Sharath Turuvekere Sreenivas for fruitful discussions.

\bibliographystyle{plainnat}
\bibliography{Bib}

\begin{thebibliography}{63}
\providecommand{\natexlab}[1]{#1}
\providecommand{\url}[1]{\texttt{#1}}
\expandafter\ifx\csname urlstyle\endcsname\relax
  \providecommand{\doi}[1]{doi: #1}\else
  \providecommand{\doi}{doi: \begingroup \urlstyle{rm}\Url}\fi

\bibitem[cla()]{claude3}
The claude 3 model family: Opus, sonnet, haiku.
\newblock URL \url{https://api.semanticscholar.org/CorpusID:268232499}.

\bibitem[Adler et~al.(2024)Adler, Agarwal, Aithal, Anh, Bhattacharya, Brundyn, Casper, Catanzaro, Clay, Cohen, Das, Dattagupta, Delalleau, Derczynski, Dong, Egert, Evans, Ficek, Fridman, Ghosh, Ginsburg, Gitman, Grzegorzek, Hero, Huang, Jawa, Jennings, Jhunjhunwala, Kamalu, Khan, Kuchaiev, LeGresley, Li, Liu, Liu, Long, Mahabaleshwarkar, Majumdar, Maki, Martinez, de~Melo, Moshkov, Narayanan, Narenthiran, Navarro, Nguyen, Nitski, Noroozi, Nutheti, Parisien, Parmar, Patwary, Pawelec, Ping, Prabhumoye, Roy, Saar, Sabavat, Satheesh, Scowcroft, Sewall, Shamis, Shen, Shoeybi, Sizer, Smelyanskiy, Soares, Sreedhar, Su, Subramanian, Sun, Toshniwal, Wang, Wang, You, Zeng, Zhang, Zhang, Zhang, Zhang, and Zhu]{Nemotron-4-340B}
Bo~Adler, Niket Agarwal, Ashwath Aithal, Dong~H. Anh, Pallab Bhattacharya, Annika Brundyn, Jared Casper, Bryan Catanzaro, Sharon Clay, Jonathan~M. Cohen, Sirshak Das, Ayush Dattagupta, Olivier Delalleau, Leon Derczynski, Yi~Dong, Daniel Egert, Ellie Evans, Aleksander Ficek, Denys Fridman, Shaona Ghosh, Boris Ginsburg, Igor Gitman, Tomasz Grzegorzek, Robert Hero, Jining Huang, Vibhu Jawa, Joseph Jennings, Aastha Jhunjhunwala, John Kamalu, Sadaf Khan, Oleksii Kuchaiev, Patrick LeGresley, Hui Li, Jiwei Liu, Zihan Liu, Eileen Long, Ameya~Sunil Mahabaleshwarkar, Somshubra Majumdar, James Maki, Miguel Martinez, Maer~Rodrigues de~Melo, Ivan Moshkov, Deepak Narayanan, Sean Narenthiran, Jesus Navarro, Phong Nguyen, Osvald Nitski, Vahid Noroozi, Guruprasad Nutheti, Christopher Parisien, Jupinder Parmar, Mostofa Patwary, Krzysztof Pawelec, Wei Ping, Shrimai Prabhumoye, Rajarshi Roy, Trisha Saar, Vasanth Rao~Naik Sabavat, Sanjeev Satheesh, Jane~Polak Scowcroft, Jason Sewall, Pavel Shamis, Gerald Shen, Mohammad Shoeybi,
  Dave Sizer, Misha Smelyanskiy, Felipe Soares, Makesh~Narsimhan Sreedhar, Dan Su, Sandeep Subramanian, Shengyang Sun, Shubham Toshniwal, Hao Wang, Zhilin Wang, Jiaxuan You, Jiaqi Zeng, Jimmy Zhang, Jing Zhang, Vivienne Zhang, Yian Zhang, and Chen Zhu.
\newblock Nemotron-4 340b technical report.
\newblock \emph{CoRR}, abs/2406.11704, 2024.
\newblock \doi{10.48550/ARXIV.2406.11704}.
\newblock URL \url{https://doi.org/10.48550/arXiv.2406.11704}.

\bibitem[Aghajanyan et~al.(2021)Aghajanyan, Gupta, and Zettlemoyer]{intrinsic-dimensionality}
Armen Aghajanyan, Sonal Gupta, and Luke Zettlemoyer.
\newblock Intrinsic dimensionality explains the effectiveness of language model fine-tuning.
\newblock In Chengqing Zong, Fei Xia, Wenjie Li, and Roberto Navigli, editors, \emph{Proceedings of the 59th Annual Meeting of the Association for Computational Linguistics and the 11th International Joint Conference on Natural Language Processing (Volume 1: Long Papers)}, pages 7319--7328, Online, August 2021. Association for Computational Linguistics.
\newblock \doi{10.18653/v1/2021.acl-long.568}.
\newblock URL \url{https://aclanthology.org/2021.acl-long.568}.

\bibitem[Ainslie et~al.(2023)Ainslie, Lee{-}Thorp, de~Jong, Zemlyanskiy, Lebr{\'{o}}n, and Sanghai]{gqa}
Joshua Ainslie, James Lee{-}Thorp, Michiel de~Jong, Yury Zemlyanskiy, Federico Lebr{\'{o}}n, and Sumit Sanghai.
\newblock {GQA:} training generalized multi-query transformer models from multi-head checkpoints.
\newblock In Houda Bouamor, Juan Pino, and Kalika Bali, editors, \emph{Proceedings of the 2023 Conference on Empirical Methods in Natural Language Processing, {EMNLP} 2023, Singapore, December 6-10, 2023}, pages 4895--4901. Association for Computational Linguistics, 2023.
\newblock \doi{10.18653/V1/2023.EMNLP-MAIN.298}.
\newblock URL \url{https://doi.org/10.18653/v1/2023.emnlp-main.298}.

\bibitem[Allen{-}Zhu et~al.(2019)Allen{-}Zhu, Li, and Song]{over-parameterization}
Zeyuan Allen{-}Zhu, Yuanzhi Li, and Zhao Song.
\newblock A convergence theory for deep learning via over-parameterization.
\newblock In Kamalika Chaudhuri and Ruslan Salakhutdinov, editors, \emph{Proceedings of the 36th International Conference on Machine Learning, {ICML} 2019, 9-15 June 2019, Long Beach, California, {USA}}, volume~97 of \emph{Proceedings of Machine Learning Research}, pages 242--252. {PMLR}, 2019.
\newblock URL \url{http://proceedings.mlr.press/v97/allen-zhu19a.html}.

\bibitem[Anil et~al.(2023)Anil, Borgeaud, Wu, Alayrac, Yu, Soricut, Schalkwyk, Dai, Hauth, Millican, Silver, Petrov, Johnson, Antonoglou, Schrittwieser, Glaese, Chen, Pitler, Lillicrap, Lazaridou, Firat, Molloy, Isard, Barham, Hennigan, Lee, Viola, Reynolds, Xu, Doherty, Collins, Meyer, Rutherford, Moreira, Ayoub, Goel, Tucker, Piqueras, Krikun, Barr, Savinov, Danihelka, Roelofs, White, Andreassen, von Glehn, Yagati, Kazemi, Gonzalez, Khalman, Sygnowski, and et~al.]{gemini}
Rohan Anil, Sebastian Borgeaud, Yonghui Wu, Jean{-}Baptiste Alayrac, Jiahui Yu, Radu Soricut, Johan Schalkwyk, Andrew~M. Dai, Anja Hauth, Katie Millican, David Silver, Slav Petrov, Melvin Johnson, Ioannis Antonoglou, Julian Schrittwieser, Amelia Glaese, Jilin Chen, Emily Pitler, Timothy~P. Lillicrap, Angeliki Lazaridou, Orhan Firat, James Molloy, Michael Isard, Paul~Ronald Barham, Tom Hennigan, Benjamin Lee, Fabio Viola, Malcolm Reynolds, Yuanzhong Xu, Ryan Doherty, Eli Collins, Clemens Meyer, Eliza Rutherford, Erica Moreira, Kareem Ayoub, Megha Goel, George Tucker, Enrique Piqueras, Maxim Krikun, Iain Barr, Nikolay Savinov, Ivo Danihelka, Becca Roelofs, Ana{\"{\i}}s White, Anders Andreassen, Tamara von Glehn, Lakshman Yagati, Mehran Kazemi, Lucas Gonzalez, Misha Khalman, Jakub Sygnowski, and et~al.
\newblock Gemini: {A} family of highly capable multimodal models.
\newblock \emph{CoRR}, abs/2312.11805, 2023.
\newblock \doi{10.48550/ARXIV.2312.11805}.
\newblock URL \url{https://doi.org/10.48550/arXiv.2312.11805}.

\bibitem[Ashkboos et~al.(2024)Ashkboos, Croci, Nascimento, Hoefler, and Hensman]{sliceGPT}
Saleh Ashkboos, Maximilian~L. Croci, Marcelo Gennari~Do Nascimento, Torsten Hoefler, and James Hensman.
\newblock Slicegpt: Compress large language models by deleting rows and columns.
\newblock In \emph{The Twelfth International Conference on Learning Representations, {ICLR} 2024, Vienna, Austria, May 7-11, 2024}. OpenReview.net, 2024.
\newblock URL \url{https://openreview.net/forum?id=vXxardq6db}.

\bibitem[Belkin et~al.(2018)Belkin, Hsu, Ma, and Mandal]{double-descent}
Mikhail Belkin, Daniel~J. Hsu, Siyuan Ma, and Soumik Mandal.
\newblock Reconciling modern machine-learning practice and the classical bias–variance trade-off.
\newblock \emph{Proceedings of the National Academy of Sciences}, 116:\penalty0 15849 -- 15854, 2018.
\newblock URL \url{https://api.semanticscholar.org/CorpusID:198496504}.

\bibitem[Beltagy et~al.(2020)Beltagy, Peters, and Cohan]{longformer_window_attention}
Iz~Beltagy, Matthew~E. Peters, and Arman Cohan.
\newblock Longformer: The long-document transformer.
\newblock \emph{CoRR}, abs/2004.05150, 2020.
\newblock URL \url{https://arxiv.org/abs/2004.05150}.

\bibitem[Bercovich et~al.(2025)Bercovich, Dabbah, Puny, Galil, Geifman, Geifman, Golan, Karpas, Levy, Moshe, Nabwani, Ronen, Schen, Segal, Shahaf, Tropp, Zilberstein, and El{-}Yaniv]{ffn_fusion}
Akhiad Bercovich, Mohammad Dabbah, Omri Puny, Ido Galil, Amnon Geifman, Yonatan Geifman, Izhak Golan, Ehud Karpas, Itay Levy, Zach Moshe, Najeeb Nabwani, Tomer Ronen, Itamar Schen, Elad Segal, Ido Shahaf, Oren Tropp, Ran Zilberstein, and Ran El{-}Yaniv.
\newblock {FFN} fusion: Rethinking sequential computation in large language models.
\newblock \emph{CoRR}, abs/2503.18908, 2025.
\newblock \doi{10.48550/ARXIV.2503.18908}.
\newblock URL \url{https://doi.org/10.48550/arXiv.2503.18908}.

\bibitem[Blakeman et~al.(2025)Blakeman, Basant, Khattar, Renduchintala, Bercovich, Ficek, Bjorlin, Taghibakhshi, Deshmukh, Mahabaleshwarkar, Tao, Shors, Aithal, Poojary, Dattagupta, Buddharaju, Chen, Ginsburg, Wang, Norick, Butterfield, Catanzaro, del Mundo, Dong, Harvey, Parisien, Su, Korzekwa, Yin, Gitman, Mosallanezhad, Narayanan, Fridman, Rekesh, Ma, Pykhtar, Ahn, Riach, Stosic, Long, Segal, Evans, Chung, Galinkin, Bakhturina, Dobrowolska, Jia, Liu, Prasad, Shen, Liu, Chen, Qian, Ngo, Liu, Li, Gitman, Karmanov, Moshkov, Golan, Kautz, Scowcroft, Casper, Sepp{\"{a}}nen, Lu, Sewall, Zeng, You, Zhang, Zhang, Huang, Xue, Huang, Conway, Kamalu, Barker, Cohen, Jennings, Parmar, Sapra, Briski, Chumachenko, Luna, Santhanam, Kong, Sivamani, Pawelec, Anik, Li, McAfee, Derczynski, Pavao, Vega, Voegtle, Bala, de~Melo, Sreedhar, Chochowski, and Kliegl]{nemotron-h}
Aaron Blakeman, Aarti Basant, Abhinav Khattar, Adithya Renduchintala, Akhiad Bercovich, Aleksander Ficek, Alexis Bjorlin, Ali Taghibakhshi, Amala~Sanjay Deshmukh, Ameya~Sunil Mahabaleshwarkar, Andrew Tao, Anna Shors, Ashwath Aithal, Ashwin Poojary, Ayush Dattagupta, Balaram Buddharaju, Bobby Chen, Boris Ginsburg, Boxin Wang, Brandon Norick, Brian Butterfield, Bryan Catanzaro, Carlo del Mundo, Chengyu Dong, Christine Harvey, Christopher Parisien, Dan Su, Daniel Korzekwa, Danny Yin, Daria Gitman, David Mosallanezhad, Deepak Narayanan, Denys Fridman, Dima Rekesh, Ding Ma, Dmytro Pykhtar, Dong Ahn, Duncan Riach, Dusan Stosic, Eileen Long, Elad Segal, Ellie Evans, Eric Chung, Erick Galinkin, Evelina Bakhturina, Ewa Dobrowolska, Fei Jia, Fuxiao Liu, Gargi Prasad, Gerald Shen, Guilin Liu, Guo Chen, Haifeng Qian, Helen Ngo, Hongbin Liu, Hui Li, Igor Gitman, Ilia Karmanov, Ivan Moshkov, Izik Golan, Jan Kautz, Jane~Polak Scowcroft, Jared Casper, Jarno Sepp{\"{a}}nen, Jason Lu, Jason Sewall, Jiaqi Zeng, Jiaxuan You,
  Jimmy Zhang, Jing Zhang, Jining Huang, Jinze Xue, Jocelyn Huang, Joey Conway, John Kamalu, Jon Barker, Jonathan~M. Cohen, Joseph Jennings, Jupinder Parmar, Karan Sapra, Kari Briski, Kateryna Chumachenko, Katherine Luna, Keshav Santhanam, Kezhi Kong, Kirthi Sivamani, Krzysztof Pawelec, Kumar Anik, Kunlun Li, Lawrence McAfee, Leon Derczynski, Lindsey Pavao, Luis Vega, Lukas Voegtle, Maciej Bala, Maer~Rodrigues de~Melo, Makesh~Narsimhan Sreedhar, Marcin Chochowski, and Markus Kliegl.
\newblock Nemotron-h: {A} family of accurate and efficient hybrid mamba-transformer models.
\newblock \emph{CoRR}, abs/2504.03624, 2025.
\newblock \doi{10.48550/ARXIV.2504.03624}.
\newblock URL \url{https://doi.org/10.48550/arXiv.2504.03624}.

\bibitem[Chen et~al.(2021)Chen, Tworek, Jun, Yuan, de~Oliveira~Pinto, Kaplan, Edwards, Burda, Joseph, Brockman, Ray, Puri, Krueger, Petrov, Khlaaf, Sastry, Mishkin, Chan, Gray, Ryder, Pavlov, Power, Kaiser, Bavarian, Winter, Tillet, Such, Cummings, Plappert, Chantzis, Barnes, Herbert{-}Voss, Guss, Nichol, Paino, Tezak, Tang, Babuschkin, Balaji, Jain, Saunders, Hesse, Carr, Leike, Achiam, Misra, Morikawa, Radford, Knight, Brundage, Murati, Mayer, Welinder, McGrew, Amodei, McCandlish, Sutskever, and Zaremba]{humaneval}
Mark Chen, Jerry Tworek, Heewoo Jun, Qiming Yuan, Henrique~Pond{\'{e}} de~Oliveira~Pinto, Jared Kaplan, Harri Edwards, Yuri Burda, Nicholas Joseph, Greg Brockman, Alex Ray, Raul Puri, Gretchen Krueger, Michael Petrov, Heidy Khlaaf, Girish Sastry, Pamela Mishkin, Brooke Chan, Scott Gray, Nick Ryder, Mikhail Pavlov, Alethea Power, Lukasz Kaiser, Mohammad Bavarian, Clemens Winter, Philippe Tillet, Felipe~Petroski Such, Dave Cummings, Matthias Plappert, Fotios Chantzis, Elizabeth Barnes, Ariel Herbert{-}Voss, William~Hebgen Guss, Alex Nichol, Alex Paino, Nikolas Tezak, Jie Tang, Igor Babuschkin, Suchir Balaji, Shantanu Jain, William Saunders, Christopher Hesse, Andrew~N. Carr, Jan Leike, Joshua Achiam, Vedant Misra, Evan Morikawa, Alec Radford, Matthew Knight, Miles Brundage, Mira Murati, Katie Mayer, Peter Welinder, Bob McGrew, Dario Amodei, Sam McCandlish, Ilya Sutskever, and Wojciech Zaremba.
\newblock Evaluating large language models trained on code.
\newblock \emph{CoRR}, abs/2107.03374, 2021.
\newblock URL \url{https://arxiv.org/abs/2107.03374}.

\bibitem[Clark et~al.(2018)Clark, Cowhey, Etzioni, Khot, Sabharwal, Schoenick, and Tafjord]{arc}
Peter Clark, Isaac Cowhey, Oren Etzioni, Tushar Khot, Ashish Sabharwal, Carissa Schoenick, and Oyvind Tafjord.
\newblock Think you have solved question answering? try arc, the {AI2} reasoning challenge.
\newblock \emph{CoRR}, abs/1803.05457, 2018.
\newblock URL \url{http://arxiv.org/abs/1803.05457}.

\bibitem[Cobbe et~al.(2021)Cobbe, Kosaraju, Bavarian, Chen, Jun, Kaiser, Plappert, Tworek, Hilton, Nakano, Hesse, and Schulman]{GSM8K}
Karl Cobbe, Vineet Kosaraju, Mohammad Bavarian, Mark Chen, Heewoo Jun, Lukasz Kaiser, Matthias Plappert, Jerry Tworek, Jacob Hilton, Reiichiro Nakano, Christopher Hesse, and John Schulman.
\newblock Training verifiers to solve math word problems.
\newblock \emph{CoRR}, abs/2110.14168, 2021.
\newblock URL \url{https://arxiv.org/abs/2110.14168}.

\bibitem[Dubey et~al.(2024)Dubey, Jauhri, Pandey, Kadian, Al{-}Dahle, Letman, Mathur, Schelten, Yang, Fan, Goyal, Hartshorn, Yang, Mitra, Sravankumar, Korenev, Hinsvark, Rao, Zhang, Rodriguez, Gregerson, Spataru, Rozi{\`{e}}re, Biron, Tang, Chern, Caucheteux, Nayak, Bi, Marra, McConnell, Keller, Touret, Wu, Wong, Ferrer, Nikolaidis, Allonsius, Song, Pintz, Livshits, Esiobu, Choudhary, Mahajan, Garcia{-}Olano, Perino, Hupkes, Lakomkin, AlBadawy, Lobanova, Dinan, Smith, Radenovic, Zhang, Synnaeve, Lee, Anderson, Nail, Mialon, Pang, Cucurell, Nguyen, Korevaar, Xu, Touvron, Zarov, Ibarra, Kloumann, Misra, Evtimov, Copet, Lee, Geffert, Vranes, Park, Mahadeokar, Shah, van~der Linde, Billock, Hong, Lee, Fu, Chi, Huang, Liu, Wang, Yu, Bitton, Spisak, Park, Rocca, Johnstun, Saxe, Jia, Alwala, Upasani, Plawiak, Li, Heafield, Stone, and et~al.]{llama3}
Abhimanyu Dubey, Abhinav Jauhri, Abhinav Pandey, Abhishek Kadian, Ahmad Al{-}Dahle, Aiesha Letman, Akhil Mathur, Alan Schelten, Amy Yang, Angela Fan, Anirudh Goyal, Anthony Hartshorn, Aobo Yang, Archi Mitra, Archie Sravankumar, Artem Korenev, Arthur Hinsvark, Arun Rao, Aston Zhang, Aur{\'{e}}lien Rodriguez, Austen Gregerson, Ava Spataru, Baptiste Rozi{\`{e}}re, Bethany Biron, Binh Tang, Bobbie Chern, Charlotte Caucheteux, Chaya Nayak, Chloe Bi, Chris Marra, Chris McConnell, Christian Keller, Christophe Touret, Chunyang Wu, Corinne Wong, Cristian~Canton Ferrer, Cyrus Nikolaidis, Damien Allonsius, Daniel Song, Danielle Pintz, Danny Livshits, David Esiobu, Dhruv Choudhary, Dhruv Mahajan, Diego Garcia{-}Olano, Diego Perino, Dieuwke Hupkes, Egor Lakomkin, Ehab AlBadawy, Elina Lobanova, Emily Dinan, Eric~Michael Smith, Filip Radenovic, Frank Zhang, Gabriel Synnaeve, Gabrielle Lee, Georgia~Lewis Anderson, Graeme Nail, Gr{\'{e}}goire Mialon, Guan Pang, Guillem Cucurell, Hailey Nguyen, Hannah Korevaar, Hu~Xu, Hugo
  Touvron, Iliyan Zarov, Imanol~Arrieta Ibarra, Isabel~M. Kloumann, Ishan Misra, Ivan Evtimov, Jade Copet, Jaewon Lee, Jan Geffert, Jana Vranes, Jason Park, Jay Mahadeokar, Jeet Shah, Jelmer van~der Linde, Jennifer Billock, Jenny Hong, Jenya Lee, Jeremy Fu, Jianfeng Chi, Jianyu Huang, Jiawen Liu, Jie Wang, Jiecao Yu, Joanna Bitton, Joe Spisak, Jongsoo Park, Joseph Rocca, Joshua Johnstun, Joshua Saxe, Junteng Jia, Kalyan~Vasuden Alwala, Kartikeya Upasani, Kate Plawiak, Ke~Li, Kenneth Heafield, Kevin Stone, and et~al.
\newblock The llama 3 herd of models.
\newblock \emph{CoRR}, abs/2407.21783, 2024.
\newblock \doi{10.48550/ARXIV.2407.21783}.
\newblock URL \url{https://doi.org/10.48550/arXiv.2407.21783}.

\bibitem[Gromov et~al.(2024)Gromov, Tirumala, Shapourian, Glorioso, and Roberts]{layer_pruning_Ineffectiveness_of_deeper_layers}
Andrey Gromov, Kushal Tirumala, Hassan Shapourian, Paolo Glorioso, and Daniel~A. Roberts.
\newblock The unreasonable ineffectiveness of the deeper layers.
\newblock \emph{CoRR}, abs/2403.17887, 2024.
\newblock \doi{10.48550/ARXIV.2403.17887}.
\newblock URL \url{https://doi.org/10.48550/arXiv.2403.17887}.

\bibitem[Gu and Dao(2024)]{mamba}
Albert Gu and Tri Dao.
\newblock Mamba: Linear-time sequence modeling with selective state spaces.
\newblock In \emph{First Conference on Language Modeling}, 2024.
\newblock URL \url{https://openreview.net/forum?id=tEYskw1VY2}.

\bibitem[Gu et~al.(2022)Gu, Goel, and Re]{ssms}
Albert Gu, Karan Goel, and Christopher Re.
\newblock Efficiently modeling long sequences with structured state spaces.
\newblock In \emph{International Conference on Learning Representations}, 2022.
\newblock URL \url{https://openreview.net/forum?id=uYLFoz1vlAC}.

\bibitem[Hardt and Sun(2024)]{test-time-training}
Moritz Hardt and Yu~Sun.
\newblock Test-time training on nearest neighbors for large language models.
\newblock In \emph{The Twelfth International Conference on Learning Representations, {ICLR} 2024, Vienna, Austria, May 7-11, 2024}. OpenReview.net, 2024.
\newblock URL \url{https://openreview.net/forum?id=CNL2bku4ra}.

\bibitem[Hasan et~al.(2021)Hasan, Bhattacharjee, Islam, Mubasshir, Li, Kang, Rahman, and Shahriyar]{XL-Sum}
Tahmid Hasan, Abhik Bhattacharjee, Md.~Saiful Islam, Kazi~Samin Mubasshir, Yuan{-}Fang Li, Yong{-}Bin Kang, M.~Sohel Rahman, and Rifat Shahriyar.
\newblock Xl-sum: Large-scale multilingual abstractive summarization for 44 languages.
\newblock In Chengqing Zong, Fei Xia, Wenjie Li, and Roberto Navigli, editors, \emph{Findings of the Association for Computational Linguistics: {ACL/IJCNLP} 2021, Online Event, August 1-6, 2021}, volume {ACL/IJCNLP} 2021 of \emph{Findings of {ACL}}, pages 4693--4703. Association for Computational Linguistics, 2021.
\newblock \doi{10.18653/V1/2021.FINDINGS-ACL.413}.
\newblock URL \url{https://doi.org/10.18653/v1/2021.findings-acl.413}.

\bibitem[He et~al.(2024)He, Sun, Shen, and Li]{no-op_What_Matters_in_Transformers}
Shwai He, Guoheng Sun, Zheyu Shen, and Ang Li.
\newblock What matters in transformers? not all attention is needed.
\newblock \emph{CoRR}, abs/2406.15786, 2024.
\newblock \doi{10.48550/ARXIV.2406.15786}.
\newblock URL \url{https://doi.org/10.48550/arXiv.2406.15786}.

\bibitem[Hendrycks et~al.(2021)Hendrycks, Burns, Basart, Zou, Mazeika, Song, and Steinhardt]{mmlu}
Dan Hendrycks, Collin Burns, Steven Basart, Andy Zou, Mantas Mazeika, Dawn Song, and Jacob Steinhardt.
\newblock Measuring massive multitask language understanding.
\newblock In \emph{9th International Conference on Learning Representations, {ICLR} 2021, Virtual Event, Austria, May 3-7, 2021}. OpenReview.net, 2021.
\newblock URL \url{https://openreview.net/forum?id=d7KBjmI3GmQ}.

\bibitem[Hinton et~al.(2015)Hinton, Vinyals, and Dean]{hinton-kd}
Geoffrey~E. Hinton, Oriol Vinyals, and Jeffrey Dean.
\newblock Distilling the knowledge in a neural network.
\newblock \emph{CoRR}, abs/1503.02531, 2015.
\newblock URL \url{http://arxiv.org/abs/1503.02531}.

\bibitem[Hive-Digital-Technologies()]{buzz}
Hive-Digital-Technologies.
\newblock \url{https://huggingface.co/datasets/H-D-T/Buzz-V1.2}.

\bibitem[Hoffmann et~al.(2024)Hoffmann, Borgeaud, Mensch, Buchatskaya, Cai, Rutherford, de~Las~Casas, Hendricks, Welbl, Clark, Hennigan, Noland, Millican, van~den Driessche, Damoc, Guy, Osindero, Simonyan, Elsen, Vinyals, Rae, and Sifre]{chinchilla}
Jordan Hoffmann, Sebastian Borgeaud, Arthur Mensch, Elena Buchatskaya, Trevor Cai, Eliza Rutherford, Diego de~Las~Casas, Lisa~Anne Hendricks, Johannes Welbl, Aidan Clark, Tom Hennigan, Eric Noland, Katie Millican, George van~den Driessche, Bogdan Damoc, Aurelia Guy, Simon Osindero, Karen Simonyan, Erich Elsen, Oriol Vinyals, Jack~W. Rae, and Laurent Sifre.
\newblock Training compute-optimal large language models.
\newblock In \emph{Proceedings of the 36th International Conference on Neural Information Processing Systems}, NIPS '22, Red Hook, NY, USA, 2024. Curran Associates Inc.
\newblock ISBN 9781713871088.

\bibitem[Hsieh et~al.(2024)Hsieh, Sun, Kriman, Acharya, Rekesh, Jia, Zhang, and Ginsburg]{DBLP:journals/corr/abs-2404-06654}
Cheng{-}Ping Hsieh, Simeng Sun, Samuel Kriman, Shantanu Acharya, Dima Rekesh, Fei Jia, Yang Zhang, and Boris Ginsburg.
\newblock {RULER:} what's the real context size of your long-context language models?
\newblock \emph{CoRR}, abs/2404.06654, 2024.
\newblock \doi{10.48550/ARXIV.2404.06654}.
\newblock URL \url{https://doi.org/10.48550/arXiv.2404.06654}.

\bibitem[Hu et~al.(2022)Hu, Shen, Wallis, Allen{-}Zhu, Li, Wang, Wang, and Chen]{lora}
Edward~J. Hu, Yelong Shen, Phillip Wallis, Zeyuan Allen{-}Zhu, Yuanzhi Li, Shean Wang, Lu~Wang, and Weizhu Chen.
\newblock Lora: Low-rank adaptation of large language models.
\newblock In \emph{The Tenth International Conference on Learning Representations, {ICLR} 2022, Virtual Event, April 25-29, 2022}. OpenReview.net, 2022.
\newblock URL \url{https://openreview.net/forum?id=nZeVKeeFYf9}.

\bibitem[Inc.(2023)]{python-mip}
COIN-OR~Foundation Inc.
\newblock Python-mip: collection of python tools for the modeling and solution of mixed-integer linear programs, 2023.
\newblock URL \url{https://github.com/coin-or/python-mip}.

\bibitem[Kaplan et~al.(2020)Kaplan, McCandlish, Henighan, Brown, Chess, Child, Gray, Radford, Wu, and Amodei]{scaling-laws}
Jared Kaplan, Sam McCandlish, Tom Henighan, Tom~B. Brown, Benjamin Chess, Rewon Child, Scott Gray, Alec Radford, Jeffrey Wu, and Dario Amodei.
\newblock Scaling laws for neural language models.
\newblock \emph{CoRR}, abs/2001.08361, 2020.
\newblock URL \url{https://arxiv.org/abs/2001.08361}.

\bibitem[Khodak et~al.(2021)Khodak, Tenenholtz, Mackey, and Fusi]{khodak2024initialization}
Mikhail Khodak, Neil~A. Tenenholtz, Lester Mackey, and Nicol{\`{o}} Fusi.
\newblock Initialization and regularization of factorized neural layers.
\newblock In \emph{9th International Conference on Learning Representations, {ICLR} 2021, Virtual Event, Austria, May 3-7, 2021}. OpenReview.net, 2021.
\newblock URL \url{https://openreview.net/forum?id=KTlJT1nof6d}.

\bibitem[Kurtic et~al.(2023)Kurtic, Kuznedelev, Frantar, Goin, and Alistarh]{square_head}
Eldar Kurtic, Denis Kuznedelev, Elias Frantar, Michael Goin, and Dan Alistarh.
\newblock Sparse fine-tuning for inference acceleration of large language models.
\newblock \emph{CoRR}, abs/2310.06927, 2023.
\newblock \doi{10.48550/ARXIV.2310.06927}.
\newblock URL \url{https://doi.org/10.48550/arXiv.2310.06927}.

\bibitem[Kwon et~al.(2023)Kwon, Li, Zhuang, Sheng, Zheng, Yu, Gonzalez, Zhang, and Stoica]{kwon2023efficientmemorymanagementlarge}
Woosuk Kwon, Zhuohan Li, Siyuan Zhuang, Ying Sheng, Lianmin Zheng, Cody~Hao Yu, Joseph~E. Gonzalez, Hao Zhang, and Ion Stoica.
\newblock Efficient memory management for large language model serving with pagedattention, 2023.
\newblock URL \url{https://arxiv.org/abs/2309.06180}.

\bibitem[Lewis et~al.(2020)Lewis, Perez, Piktus, Petroni, Karpukhin, Goyal, Kuttler, Lewis, tau Yih, Rockt{\"a}schel, Riedel, and Kiela]{rag}
Patrick Lewis, Ethan Perez, Aleksandara Piktus, Fabio Petroni, Vladimir Karpukhin, Naman Goyal, Heinrich Kuttler, Mike Lewis, Wen tau Yih, Tim Rockt{\"a}schel, Sebastian Riedel, and Douwe Kiela.
\newblock Retrieval-augmented generation for knowledge-intensive nlp tasks.
\newblock \emph{ArXiv}, abs/2005.11401, 2020.
\newblock URL \url{https://api.semanticscholar.org/CorpusID:218869575}.

\bibitem[Li et~al.(2019)Li, Peng, Yuan, Wang, Liang, Lin, and Chang]{dna}
Changlin Li, Jiefeng Peng, Liuchun Yuan, Guangrun Wang, Xiaodan Liang, Liang Lin, and Xiaojun Chang.
\newblock Block-wisely supervised neural architecture search with knowledge distillation.
\newblock \emph{2020 IEEE/CVF Conference on Computer Vision and Pattern Recognition (CVPR)}, pages 1986--1995, 2019.
\newblock URL \url{https://api.semanticscholar.org/CorpusID:208513081}.

\bibitem[Li et~al.(2024)Li, Chiang, Frick, Dunlap, Wu, Zhu, Gonzalez, and Stoica]{arena_hard}
Tianle Li, Wei{-}Lin Chiang, Evan Frick, Lisa Dunlap, Tianhao Wu, Banghua Zhu, Joseph~E. Gonzalez, and Ion Stoica.
\newblock From crowdsourced data to high-quality benchmarks: Arena-hard and benchbuilder pipeline.
\newblock \emph{CoRR}, abs/2406.11939, 2024.
\newblock \doi{10.48550/ARXIV.2406.11939}.
\newblock URL \url{https://doi.org/10.48550/arXiv.2406.11939}.

\bibitem[Lin et~al.(2022)Lin, Hilton, and Evans]{TruthfulQA}
Stephanie Lin, Jacob Hilton, and Owain Evans.
\newblock Truthfulqa: Measuring how models mimic human falsehoods.
\newblock In Smaranda Muresan, Preslav Nakov, and Aline Villavicencio, editors, \emph{Proceedings of the 60th Annual Meeting of the Association for Computational Linguistics (Volume 1: Long Papers), {ACL} 2022, Dublin, Ireland, May 22-27, 2022}, pages 3214--3252. Association for Computational Linguistics, 2022.
\newblock \doi{10.18653/V1/2022.ACL-LONG.229}.
\newblock URL \url{https://doi.org/10.18653/v1/2022.acl-long.229}.

\bibitem[Ling et~al.(2024)Ling, Wang, Yan, and Liu]{slimgpt}
Gui Ling, Ziyang Wang, Yuliang Yan, and Qingwen Liu.
\newblock Slimgpt: Layer-wise structured pruning for large language models.
\newblock In Amir Globersons, Lester Mackey, Danielle Belgrave, Angela Fan, Ulrich Paquet, Jakub~M. Tomczak, and Cheng Zhang, editors, \emph{Advances in Neural Information Processing Systems 38: Annual Conference on Neural Information Processing Systems 2024, NeurIPS 2024, Vancouver, BC, Canada, December 10 - 15, 2024}, 2024.
\newblock URL \url{http://papers.nips.cc/paper\_files/paper/2024/hash/c1c44e46358e0fb94dc94ec495a7fb1a-Abstract-Conference.html}.

\bibitem[Liu et~al.(2023)Liu, Li, Wu, and Lee]{llava}
Haotian Liu, Chunyuan Li, Qingyang Wu, and Yong~Jae Lee.
\newblock Visual instruction tuning.
\newblock \emph{ArXiv}, abs/2304.08485, 2023.
\newblock URL \url{https://api.semanticscholar.org/CorpusID:258179774}.

\bibitem[Lu et~al.(2022)Lu, Zhang, Chu, Chen, Zhou, Wu, Chen, and Yang]{lu2022knowledge}
Chengqiang Lu, Jianwei Zhang, Yunfei Chu, Zhengyu Chen, Jingren Zhou, Fei Wu, Haiqing Chen, and Hongxia Yang.
\newblock Knowledge distillation of transformer-based language models revisited.
\newblock \emph{arXiv preprint arXiv:2206.14366}, 2022.

\bibitem[Ma et~al.(2023)Ma, Fang, and Wang]{LLM-Pruner}
Xinyin Ma, Gongfan Fang, and Xinchao Wang.
\newblock Llm-pruner: On the structural pruning of large language models.
\newblock In Alice Oh, Tristan Naumann, Amir Globerson, Kate Saenko, Moritz Hardt, and Sergey Levine, editors, \emph{Advances in Neural Information Processing Systems 36: Annual Conference on Neural Information Processing Systems 2023, NeurIPS 2023, New Orleans, LA, USA, December 10 - 16, 2023}, 2023.

\bibitem[Men et~al.(2024)Men, Xu, Zhang, Wang, Lin, Lu, Han, and Chen]{shortGPT}
Xin Men, Mingyu Xu, Qingyu Zhang, Bingning Wang, Hongyu Lin, Yaojie Lu, Xianpei Han, and Weipeng Chen.
\newblock Shortgpt: Layers in large language models are more redundant than you expect.
\newblock \emph{CoRR}, abs/2403.03853, 2024.
\newblock \doi{10.48550/ARXIV.2403.03853}.
\newblock URL \url{https://doi.org/10.48550/arXiv.2403.03853}.

\bibitem[Molchanov et~al.(2022)Molchanov, Hall, Yin, Kautz, Fusi, and Vahdat]{lana}
Pavlo Molchanov, Jimmy Hall, Hongxu Yin, Jan Kautz, Nicol{\`{o}} Fusi, and Arash Vahdat.
\newblock {LANA:} latency aware network acceleration.
\newblock In Shai Avidan, Gabriel~J. Brostow, Moustapha Ciss{\'{e}}, Giovanni~Maria Farinella, and Tal Hassner, editors, \emph{Computer Vision - {ECCV} 2022 - 17th European Conference, Tel Aviv, Israel, October 23-27, 2022, Proceedings, Part {XII}}, volume 13672 of \emph{Lecture Notes in Computer Science}, pages 137--156. Springer, 2022.
\newblock \doi{10.1007/978-3-031-19775-8\_9}.
\newblock URL \url{https://doi.org/10.1007/978-3-031-19775-8\_9}.

\bibitem[Moons et~al.(2020)Moons, Noorzad, Skliar, Mariani, Mehta, Lott, and Blankevoort]{donna}
Bert Moons, Parham Noorzad, Andrii Skliar, Giovanni Mariani, Dushyant Mehta, Chris Lott, and Tijmen Blankevoort.
\newblock Distilling optimal neural networks: Rapid search in diverse spaces.
\newblock \emph{2021 IEEE/CVF International Conference on Computer Vision (ICCV)}, pages 12209--12218, 2020.
\newblock URL \url{https://api.semanticscholar.org/CorpusID:229211047}.

\bibitem[Muralidharan et~al.(2024)Muralidharan, Sreenivas, Joshi, Chochowski, Patwary, Shoeybi, Catanzaro, Kautz, and Molchanov]{minitron_Compact_Language_Models_via_Pruning_and_Knowledge_Distillation}
Saurav Muralidharan, Sharath~Turuvekere Sreenivas, Raviraj Joshi, Marcin Chochowski, Mostofa Patwary, Mohammad Shoeybi, Bryan Catanzaro, Jan Kautz, and Pavlo Molchanov.
\newblock Compact language models via pruning and knowledge distillation.
\newblock \emph{CoRR}, abs/2407.14679, 2024.
\newblock \doi{10.48550/ARXIV.2407.14679}.
\newblock URL \url{https://doi.org/10.48550/arXiv.2407.14679}.

\bibitem[OpenAI(2023)]{gpt4}
OpenAI.
\newblock {GPT-4} technical report.
\newblock \emph{CoRR}, abs/2303.08774, 2023.
\newblock \doi{10.48550/ARXIV.2303.08774}.
\newblock URL \url{https://doi.org/10.48550/arXiv.2303.08774}.

\bibitem[Penedo et~al.(2024)Penedo, Kydl{\'{\i}}cek, Allal, Lozhkov, Mitchell, Raffel, von Werra, and Wolf]{fineweb}
Guilherme Penedo, Hynek Kydl{\'{\i}}cek, Loubna~Ben Allal, Anton Lozhkov, Margaret Mitchell, Colin Raffel, Leandro von Werra, and Thomas Wolf.
\newblock The fineweb datasets: Decanting the web for the finest text data at scale.
\newblock \emph{CoRR}, abs/2406.17557, 2024.
\newblock \doi{10.48550/ARXIV.2406.17557}.
\newblock URL \url{https://doi.org/10.48550/arXiv.2406.17557}.

\bibitem[{Project Gutenberg}()]{projectgutenberg}
{Project Gutenberg}.
\newblock \url{https://www.gutenberg.org}.

\bibitem[Real et~al.(2017)Real, Moore, Selle, Saxena, Suematsu, Tan, Le, and Kurakin]{evolutionary_nas}
Esteban Real, Sherry Moore, Andrew Selle, Saurabh Saxena, Yutaka~Leon Suematsu, Jie Tan, Quoc~V. Le, and Alexey Kurakin.
\newblock Large-scale evolution of image classifiers.
\newblock In \emph{International Conference on Machine Learning}, 2017.
\newblock URL \url{https://api.semanticscholar.org/CorpusID:743641}.

\bibitem[Sakaguchi et~al.(2020)Sakaguchi, Bras, Bhagavatula, and Choi]{Winogrande}
Keisuke Sakaguchi, Ronan~Le Bras, Chandra Bhagavatula, and Yejin Choi.
\newblock Winogrande: An adversarial winograd schema challenge at scale.
\newblock In \emph{The Thirty-Fourth {AAAI} Conference on Artificial Intelligence, {AAAI} 2020, The Thirty-Second Innovative Applications of Artificial Intelligence Conference, {IAAI} 2020, The Tenth {AAAI} Symposium on Educational Advances in Artificial Intelligence, {EAAI} 2020, New York, NY, USA, February 7-12, 2020}, pages 8732--8740. {AAAI} Press, 2020.
\newblock \doi{10.1609/AAAI.V34I05.6399}.
\newblock URL \url{https://doi.org/10.1609/aaai.v34i05.6399}.

\bibitem[Sanyal et~al.(2024)Sanyal, Shwartz-Ziv, Dimakis, and Sanghavi]{Inheritune}
Sunny Sanyal, Ravid Shwartz-Ziv, Alexandros~G. Dimakis, and Sujay Sanghavi.
\newblock Inheritune: Training smaller yet more attentive language models, 2024.
\newblock URL \url{https://arxiv.org/abs/2404.08634}.

\bibitem[Shazeer(2019)]{mqa}
Noam Shazeer.
\newblock Fast transformer decoding: One write-head is all you need.
\newblock \emph{CoRR}, abs/1911.02150, 2019.
\newblock URL \url{http://arxiv.org/abs/1911.02150}.

\bibitem[Soldaini et~al.(2024)Soldaini, Kinney, Bhagia, Schwenk, Atkinson, Authur, Bogin, Chandu, Dumas, Elazar, Hofmann, Jha, Kumar, Lucy, Lyu, Lambert, Magnusson, Morrison, Muennighoff, Naik, Nam, Peters, Ravichander, Richardson, Shen, Strubell, Subramani, Tafjord, Walsh, Zettlemoyer, Smith, Hajishirzi, Beltagy, Groeneveld, Dodge, and Lo]{dolma}
Luca Soldaini, Rodney Kinney, Akshita Bhagia, Dustin Schwenk, David Atkinson, Russell Authur, Ben Bogin, Khyathi~Raghavi Chandu, Jennifer Dumas, Yanai Elazar, Valentin Hofmann, Ananya~Harsh Jha, Sachin Kumar, Li~Lucy, Xinxi Lyu, Nathan Lambert, Ian Magnusson, Jacob Morrison, Niklas Muennighoff, Aakanksha Naik, Crystal Nam, Matthew~E. Peters, Abhilasha Ravichander, Kyle Richardson, Zejiang Shen, Emma Strubell, Nishant Subramani, Oyvind Tafjord, Evan~Pete Walsh, Luke Zettlemoyer, Noah~A. Smith, Hannaneh Hajishirzi, Iz~Beltagy, Dirk Groeneveld, Jesse Dodge, and Kyle Lo.
\newblock Dolma: an open corpus of three trillion tokens for language model pretraining research.
\newblock In Lun{-}Wei Ku, Andre Martins, and Vivek Srikumar, editors, \emph{Proceedings of the 62nd Annual Meeting of the Association for Computational Linguistics (Volume 1: Long Papers), {ACL} 2024, Bangkok, Thailand, August 11-16, 2024}, pages 15725--15788. Association for Computational Linguistics, 2024.
\newblock \doi{10.18653/V1/2024.ACL-LONG.840}.
\newblock URL \url{https://doi.org/10.18653/v1/2024.acl-long.840}.

\bibitem[Sun et~al.(2024)Sun, Liu, Bair, and Kolter]{sun2023wanda}
Mingjie Sun, Zhuang Liu, Anna Bair, and J.~Zico Kolter.
\newblock A simple and effective pruning approach for large language models.
\newblock In \emph{The Twelfth International Conference on Learning Representations, {ICLR} 2024, Vienna, Austria, May 7-11, 2024}. OpenReview.net, 2024.
\newblock URL \url{https://openreview.net/forum?id=PxoFut3dWW}.

\bibitem[Wang et~al.(2024)Wang, Bukharin, Delalleau, Egert, Shen, Zeng, Kuchaiev, and Dong]{HelpSteer2}
Zhilin Wang, Alexander Bukharin, Olivier Delalleau, Daniel Egert, Gerald Shen, Jiaqi Zeng, Oleksii Kuchaiev, and Yi~Dong.
\newblock Helpsteer2-preference: Complementing ratings with preferences.
\newblock \emph{CoRR}, abs/2410.01257, 2024.
\newblock \doi{10.48550/ARXIV.2410.01257}.
\newblock URL \url{https://doi.org/10.48550/arXiv.2410.01257}.

\bibitem[Wei et~al.(2022)Wei, Wang, Schuurmans, Bosma, Ichter, Xia, Chi, Le, and Zhou]{chain-of-thought}
Jason Wei, Xuezhi Wang, Dale Schuurmans, Maarten Bosma, Brian Ichter, Fei Xia, Ed~H. Chi, Quoc~V. Le, and Denny Zhou.
\newblock Chain-of-thought prompting elicits reasoning in large language models.
\newblock In Sanmi Koyejo, S.~Mohamed, A.~Agarwal, Danielle Belgrave, K.~Cho, and A.~Oh, editors, \emph{Advances in Neural Information Processing Systems 35: Annual Conference on Neural Information Processing Systems 2022, NeurIPS 2022, New Orleans, LA, USA, November 28 - December 9, 2022}, 2022.

\bibitem[Xia et~al.(2022)Xia, Zhong, and Chen]{CoFiPruning}
Mengzhou Xia, Zexuan Zhong, and Danqi Chen.
\newblock Structured pruning learns compact and accurate models.
\newblock In Smaranda Muresan, Preslav Nakov, and Aline Villavicencio, editors, \emph{Proceedings of the 60th Annual Meeting of the Association for Computational Linguistics (Volume 1: Long Papers), {ACL} 2022, Dublin, Ireland, May 22-27, 2022}, pages 1513--1528. Association for Computational Linguistics, 2022.
\newblock \doi{10.18653/V1/2022.ACL-LONG.107}.
\newblock URL \url{https://doi.org/10.18653/v1/2022.acl-long.107}.

\bibitem[Xia et~al.(2024)Xia, Gao, Zeng, and Chen]{sheared-llama}
Mengzhou Xia, Tianyu Gao, Zhiyuan Zeng, and Danqi Chen.
\newblock Sheared llama: Accelerating language model pre-training via structured pruning.
\newblock In \emph{The Twelfth International Conference on Learning Representations, {ICLR} 2024, Vienna, Austria, May 7-11, 2024}. OpenReview.net, 2024.
\newblock URL \url{https://openreview.net/forum?id=09iOdaeOzp}.

\bibitem[Xiao et~al.(2024)Xiao, Tian, Chen, Han, and Lewis]{sink_attention_streaming_llm}
Guangxuan Xiao, Yuandong Tian, Beidi Chen, Song Han, and Mike Lewis.
\newblock Efficient streaming language models with attention sinks.
\newblock In \emph{The Twelfth International Conference on Learning Representations, {ICLR} 2024, Vienna, Austria, May 7-11, 2024}. OpenReview.net, 2024.
\newblock URL \url{https://openreview.net/forum?id=NG7sS51zVF}.

\bibitem[Yao et~al.(2023)Yao, Yu, Zhao, Shafran, Griffiths, Cao, and Narasimhan]{tree-of-thoughts}
Shunyu Yao, Dian Yu, Jeffrey Zhao, Izhak Shafran, Tom Griffiths, Yuan Cao, and Karthik Narasimhan.
\newblock Tree of thoughts: Deliberate problem solving with large language models.
\newblock In Alice Oh, Tristan Naumann, Amir Globerson, Kate Saenko, Moritz Hardt, and Sergey Levine, editors, \emph{Advances in Neural Information Processing Systems 36: Annual Conference on Neural Information Processing Systems 2023, NeurIPS 2023, New Orleans, LA, USA, December 10 - 16, 2023}, 2023.

\bibitem[Zellers et~al.(2019)Zellers, Holtzman, Bisk, Farhadi, and Choi]{hellaswag}
Rowan Zellers, Ari Holtzman, Yonatan Bisk, Ali Farhadi, and Yejin Choi.
\newblock Hellaswag: Can a machine really finish your sentence?
\newblock In Anna Korhonen, David~R. Traum, and Llu{\'{\i}}s M{\`{a}}rquez, editors, \emph{Proceedings of the 57th Conference of the Association for Computational Linguistics, {ACL} 2019, Florence, Italy, July 28- August 2, 2019, Volume 1: Long Papers}, pages 4791--4800. Association for Computational Linguistics, 2019.
\newblock \doi{10.18653/V1/P19-1472}.
\newblock URL \url{https://doi.org/10.18653/v1/p19-1472}.

\bibitem[Zheng et~al.(2023)Zheng, Chiang, Sheng, Zhuang, Wu, Zhuang, Lin, Li, Li, Xing, Zhang, Gonzalez, and Stoica]{MT-bench}
Lianmin Zheng, Wei{-}Lin Chiang, Ying Sheng, Siyuan Zhuang, Zhanghao Wu, Yonghao Zhuang, Zi~Lin, Zhuohan Li, Dacheng Li, Eric~P. Xing, Hao Zhang, Joseph~E. Gonzalez, and Ion Stoica.
\newblock Judging llm-as-a-judge with mt-bench and chatbot arena.
\newblock In Alice Oh, Tristan Naumann, Amir Globerson, Kate Saenko, Moritz Hardt, and Sergey Levine, editors, \emph{Advances in Neural Information Processing Systems 36: Annual Conference on Neural Information Processing Systems 2023, NeurIPS 2023, New Orleans, LA, USA, December 10 - 16, 2023}, 2023.

\bibitem[Zhou et~al.(2024)Zhou, Yu, Babu, Tirumala, Yasunaga, Shamis, Kahn, Ma, Zettlemoyer, and Levy]{transfusion}
Chunting Zhou, Lili Yu, Arun Babu, Kushal Tirumala, Michihiro Yasunaga, Leonid Shamis, Jacob Kahn, Xuezhe Ma, Luke Zettlemoyer, and Omer Levy.
\newblock Transfusion: Predict the next token and diffuse images with one multi-modal model.
\newblock \emph{CoRR}, abs/2408.11039, 2024.
\newblock \doi{10.48550/ARXIV.2408.11039}.
\newblock URL \url{https://doi.org/10.48550/arXiv.2408.11039}.

\bibitem[Zoph and Le(2016)]{rl_nas}
Barret Zoph and Quoc~V. Le.
\newblock Neural architecture search with reinforcement learning.
\newblock \emph{ArXiv}, abs/1611.01578, 2016.
\newblock URL \url{https://api.semanticscholar.org/CorpusID:12713052}.

\end{thebibliography}

\appendix
\section{Appendices} 
\subsection{Human Evaluation} \label{sec:human_eval}
A blind-test comparison between Nemotron-51B and Llama-3.1-70B-Instruct was conducted in the following manner:

\begin{itemize}
\item A set of 169 samples was sent.
\item The evaluation was done by Nvidia's data factory team.
\item Three annotators annotated each sample independently, resulting in a total of $169\cdot3=507$ annotations.
\item Annotators saw [prompt, completion 1, completion 2] and had to choose between 4 options:
\begin{itemize}
    \item Completion 1 is better.
    \item Completion 2 is better.
    \item Both are good.
    \item Neither is good. 
\end{itemize}

\item The order of the completions was randomized to avoid positional bias.
\end{itemize}

The test set was curated by the project's product team and Nvidia's data factory and included the following tasks and subtasks listed in brackets:
\begin{itemize}
    \item Long form text generation (write a blog, write a story, other).
    \item Long inputs (Open book QA, Multi-hop questions, text to headline)
    \item Grounded text generation (table of contents to blog, paraphrasing).
    \item Multi-condition instructions (3-5 conditions, 6-10 conditions).
    \item Knowledge and trivia.
    \item Summarization (full document summary, summarize to bullet points, summarize paragraph to a sentence).
    \item Reasoning (temporal reasoning, cause and effect, navigation, general reasoning).
    \item Semantic extraction.
\end{itemize}

\begin{table}[h!]
\centering
\caption{Full performance comparison of the parent (Llama-3.1-70B-Instruct) and child (Nemotron-51B) models on a subset of the RULER benchmark across all context lengths. Accuracy preserved is the ratio of the child score to the parent score, expressed as a percentage. The names of the benchmarks refer to their implementation and settings used in the official github repository.
*Varies depending on context length}
\label{tab:full_student_teacher_performance}
\resizebox{\textwidth}{!}{
\rowcolors{2}{white}{gray!10}
\begin{tabular}{@{}lcccccccccc@{}}
\toprule
\textbf{Context Length} & \textbf{qa\_hotpotqa} & \textbf{qa\_squad} & \textbf{common\_words\_extraction*} & \textbf{variable\_tracking\_1\_4} & \textbf{variable\_tracking\_2\_2} & \textbf{freq\_words\_extraction\_2} & \textbf{freq\_words\_extraction\_3.5} & \textbf{Average} & \textbf{Accuracy Preserved (\%)} \\ \midrule
\multicolumn{10}{@{}l@{}}{\textbf{Parent}} \\
1024 & N/A & N/A & 100.00 & 100.00 & 100.00 & 99.40 & 99.67 & 99.81 & - \\
2048 & N/A & 88.40 & 100.00 & 100.00 & 100.00 & 99.53 & 99.87 & 97.97 & - \\
4096 & 67.60 & 87.40 & 100.00 & 100.00 & 99.87 & 99.00 & 99.80 & 93.38 & - \\
8192 & 67.80 & 83.80 & 99.96 & 100.00 & 99.40 & 97.87 & 99.93 & 92.68 & - \\
16384 & 63.20 & 82.00 & 98.86 & 100.00 & 96.87 & 96.67 & 99.93 & 91.08 & - \\
32768 & 61.60 & 77.20 & 93.48 & 100.00 & 97.93 & 95.53 & 100.00 & 89.39 & - \\
65536 & 55.4 & 72.60 & 26.16 & 99.96 & 97.93 & 94.53 & 99.87 & 78.06 & - \\
131072 & 33.65 & 49.04 & 2.37 & 56.85 & 36.33 & 78.61 & 85.71 & 48.94 & - \\ \midrule
\multicolumn{10}{@{}l@{}}{\textbf{Child}} \\
1024 & N/A & N/A & 99.98 & 100.00 & 100.00 & 99.40 & 99.53 & 99.78 & \textbf{99.9} \\
2048 & N/A & 86.20 & 99.86 & 99.96 & 99.67 & 98.40 & 99.80 & 97.32 & \textbf{99.34} \\
4096 & 63.40 & 85.00 & 99.92 & 100.00 & 98.93 & 97.73 & 99.87 & 92.12 & \textbf{98.65} \\
8192 & 58.20 & 80.80 & 99.34 & 100.00 & 99.60 & 96.67 & 99.80 & 90.63 & \textbf{97.79} \\
16384 & 53.40 & 75.60 & 93.50 & 99.72 & 96.80 & 94.73 & 99.80 & 87.65 & \textbf{96.23} \\
32768 & 45.60 & 70.60 & 51.92 & 98.28 & 93.67 & 90.27 & 99.47 & 78.54 & \textbf{87.86}   \\
65536 & 7.4 & 15.20 & 2.28 & 3.48 & 7.87 & 36.93 & 8.67 & 11.6 & \textbf{14.86} \\
131072 & 3.80 & 3.20 & 0.10 & 0.00 & 0.00 & 2.07 & 0.00 & 1.31 & \textbf{2.67} \\ \midrule
\end{tabular}
}
\end{table}

\section{RULER Benchmark Performance Tables}
\label{app:full_tables}

This section provides complete performance tables for our parent-child pairs on a subset of the RULER benchmark across all context lengths evaluated.

\subsection{Nemotron-51B vs.\ Llama-3.1-70B-Instruct}
Table~\ref{tab:full_student_teacher_performance} shows the results of the parent (Llama-3.1-70B-Instruct) and 
child (Nemotron-51B) for all context lengths. 
As noted in the main text, Nemotron-51B was trained on sequences up to only 8K tokens 
yet retains more than 96\% of its parent's performance at 16K. 
The child’s performance degrades at 64K and beyond, which is unsurprising given its training horizon.
Nevertheless, these results underscore that a large fraction of the parent's long-context capabilities can remain intact even without explicit training on such long sequences.

\subsection{Nemotron-49B-Base vs.\ Llama-3.3-70B-Instruct}
In Table~\ref{tab:ruler_49b}, we present the extended context-length results for Nemotron-49B-Base 
(uptrained with sequences up to 128K tokens) alongside its parent (Llama-3.3-70B-Instruct). 
We rename columns to make them more consistent with the style above, 
although some tasks differ from those used for Nemotron-51B.
Nemotron-49B-Base preserves 98\% or more of its parent’s performance up to 64K tokens 
and remains above 94\% at 128K. 
This highlights that adding a short uptraining phase on longer contexts 
can effectively extend the context range of Puzzle-derived models.
Additional details and per-task results appear below.

\begin{table}[t]
\centering
\caption{Performance comparison of the parent (Llama-3.3-70B-Instruct) and child (Nemotron-49B-Base) 
on a subset of the RULER benchmark.
`Accuracy Preserved' is (child / parent)$\times$100.}
\label{tab:ruler_49b}
\rowcolors{2}{white}{gray!10}
\resizebox{\linewidth}{!}{%
\begin{tabular}{@{}lcccccccccccccccc@{}}
\toprule
\textbf{Context} 
& \textbf{single\_needle\_1} 
& \textbf{single\_needle\_2} 
& \textbf{single\_needle\_3} 
& \textbf{multi\_needle\_1} 
& \textbf{multi\_needle\_2} 
& \textbf{multi\_needle\_3} 
& \textbf{multi\_value} 
& \textbf{multi\_query} 
& \textbf{variable\_tracking\_1\_4} 
& \textbf{common\_words\_extraction} 
& \textbf{freq\_words\_extraction} 
& \textbf{qa\_squad} 
& \textbf{qa\_hotpotqa} 
& \textbf{Average} 
& \textbf{Accuracy Preserved (\%)} \\
\midrule
\multicolumn{16}{@{}l@{}}{\textbf{Parent: Llama-3.3-70B-Instruct}} \\
4K    & 100 & 100 & 100 & 100 & 100 & 100 & 100 & 100 & 100  & 95   & 91   & 72  & 96.77 & 96.77 & - \\
8K    & 100 & 100 & 100 & 100 & 100 & 100 & 100 & 100 & 100  & 96   & 87   & 71  & 96.46 & 96.46 & - \\
16K   & 100 & 100 & 100 & 100 & 100 & 100 & 100 & 100 & 99.8 & 95   & 86   & 67  & 95.98 & 95.98 & - \\
32K   & 100 & 100 & 100 & 100 & 100 & 99  & 98.75 & 100 & 95  & 92.33 & 83 & 63  & 94.70 & 94.70 & - \\
64K   & 100 & 100 & 100 & 100 & 97  & 93  & 98.25 & 100 & 43.9 & 92.67 & 75 & 56  & 88.91 & 88.91 & - \\
128K  & 37  & 100 & 100 & 90  & 0   & 1   & 98.25 & 95.75 & 0.4 & 3.5  & 78.33 & 41 & 34 & 52.25 & 52.25 & - \\
\midrule
\multicolumn{16}{@{}l@{}}{\textbf{Child: Nemotron-49B-Base}} \\
4K    & 100 & 100 & 100 & 100 & 100 & 100 & 100 & 100 & 100  & 95.9 & 99.33 & 93  & 78 & 97.40 & \textbf{100.65} \\
8K    & 100 & 100 & 100 & 100 & 100 & 100 & 100 & 100 & 100  & 92.4 & 98.33 & 89  & 76 & 96.59 & \textbf{100.13} \\
16K   & 100 & 100 & 100 & 100 & 99  & 100 & 99.75 & 100 & 100  & 92.4 & 98   & 89  & 71 & 96.09 & \textbf{100.11} \\
32K   & 100 & 100 & 100 & 100 & 99  & 99  & 99.5  & 100 & 81.4 & 95   & 85   & 67  & 94.3 & 94.30 & \textbf{99.58} \\
64K   & 100 & 100 & 100 & 100 & 92  & 92  & 99.75 & 100 & 26.9 & 92.67 & 73   & 60  & 87.39 & 87.39 & \textbf{98.29} \\
128K  & 45  & 99  & 99  & 89  & 0   & 0   & 88    & 86.5 & 1.6 & 1.3  & 56.67 & 44 & 35 & 49.62 & \textbf{94.97} \\
\bottomrule
\end{tabular}
}
\end{table}

\end{document}